\documentclass[10pt,twocolumn,letterpaper]{article}

\usepackage[pagenumbers]{cvpr} 

\usepackage{algorithm}
\usepackage{algpseudocode}
\usepackage{makecell}
\usepackage{multirow}
\usepackage{appendix}
\usepackage{array,booktabs,multirow,colortbl,xcolor}
\usepackage{bbm}

\algrenewcommand\algorithmicrequire{\textbf{Input:}}
\algrenewcommand\algorithmicensure{\textbf{Output:}}
\usepackage[symbol]{footmisc}

\definecolor{cvprblue}{rgb}{0.21,0.49,0.74}
\usepackage[pagebackref,breaklinks,colorlinks,allcolors=cvprblue]{hyperref}

\def\confName{CVPR}
\def\confYear{2026}

\title{SwiftTailor: Efficient 3D Garment Generation with \\Geometry Image Representation}

\definecolor{mydarkblue}{rgb}{0,0.08,1}
\definecolor{mydarkgreen}{rgb}{0.02,0.6,0.02}
\definecolor{myred}{rgb}{1.0,0.0,0.0}
\definecolor{myred2}{rgb}{0.7,0.1,0.1}
\definecolor{mydarkblue2}{rgb}{0.05,0.1,0.7}
\definecolor{mypurple}{rgb}{111,0,255}
\definecolor{mypurple2}{rgb}{111,0,111}

\author{
Phuc Pham\footnotemark[1]
\and
Uy Dieu Tran\footnotemark[1] \footnotemark[2]
\and
Binh-Son Hua \footnotemark[3]
\and
Phong Nguyen
\and
\texttt{\{phucpham, phongnh\}@qti.qualcomm.com} \\
Qualcomm AI Research\footnotemark[4]
}

\begin{document}

\twocolumn[{%
\renewcommand\twocolumn[1][]{#1}%
\maketitle
\vspace{-20pt}
\begin{center}
    \includegraphics[width=\linewidth]{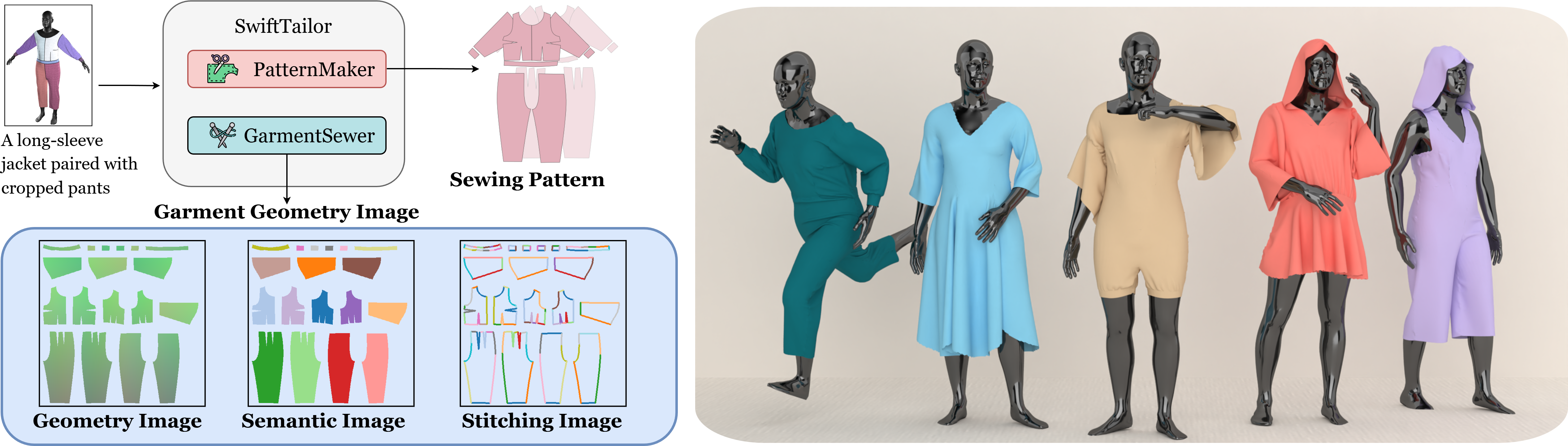}
    \captionof{figure}{We introduce SwiftTailor, a two-stage framework including PatternMaker and GarmentSewer that aims to produce sewing patterns along with a novel garment geometry image representation that can be directly decoded to final 3D garment meshes.}
    \label{fig:teaser}
\end{center}
}]

\footnotetext[1]{Equal Contribution.}
\footnotetext[2]{This work was done during the AI residency program at Qualcomm.}
\footnotetext[3]{Binh-Son Hua is affiliated with Trinity College Dublin, Ireland. Work done under consultancy capacity.}
\footnotetext[4]{Qualcomm AI Research is an initiative of Qualcomm Technologies, Inc.}

\begin{abstract}
Realistic and efficient 3D garment generation remains a longstanding challenge in computer vision and digital fashion. Existing methods typically rely on large vision- language models to produce serialized representations of 2D sewing patterns, which are then transformed into simulation-ready 3D meshes using garment modeling framework such as GarmentCode. Although these approaches yield high-quality results, they often suffer from slow inference times, ranging from 30 seconds to a minute.
In this work, we introduce SwiftTailor, a novel two-stage framework that unifies sewing-pattern reasoning and geometry-based mesh synthesis through a compact geometry image representation. SwiftTailor comprises two lightweight modules: PatternMaker, an efficient vision-language model that predicts sewing patterns from diverse input modalities, and GarmentSewer, an efficient dense prediction transformer that converts these patterns into a novel Garment Geometry Image, encoding the 3D surface of all garment panels in a unified UV space.
The final 3D mesh is reconstructed through an efficient inverse mapping process that incorporates remeshing and dynamic stitching algorithms to directly assemble the garment, thereby amortizing the cost of physical simulation. Extensive experiments on the Multimodal GarmentCodeData demonstrate that SwiftTailor achieves state-of-the-art accuracy and visual fidelity while significantly reducing inference time. This work offers a scalable, interpretable, and high-performance solution for next-generation 3D garment generation.

\end{abstract}

\section{Introduction}
Realistic and efficient 3D garment generation has long been a challenging problem in computer vision and digital fashion. 
While recent advances in general-purpose 3D generative models~\cite{atlasnet,yan2024omages64,gimdiffusion,hunyuan3d22025tencent,yang2024hunyuan3d,lai2025hunyuan3d25highfidelity3d,xiang2025structured} 
have enabled the synthesis of complex shapes, they often fail to capture the structural topology and physical realism required for clothing. 
Such models typically ignore the garment manufacturing process, resulting in meshes that are either topologically inconsistent, physically unstable under simulation, or incompatible with industrial digital-fashion workflows. 
Consequently, they fall short of the requirements of the fashion industry, where interpretability, manufacturability, and physical plausibility are crucial \cite{garmagenet}.

To bridge this gap, recent works have adopted industry-inspired workflows that first design 2D sewing patterns and then reconstruct 3D garments via physics-based sewing simulation~\cite{gc,gcd,oldgcd,chatgarment,aipparel,sewingldm,sewformer,he2024dresscode,de2023drapenet,neuralsewingmachine,neuraltailor}. 
This paradigm introduces interpretable intermediate representations and aligns well with CAD and virtual try-on systems. 
However, these approaches face several limitations. 
First, previous works ~\cite{aipparel,chatgarment} often rely on an large vision-language model(VLMs) such as  LLaVA-1.5V-7B~\cite{liu2023llava} with a CLIP~\cite{radford2021learning} visual encoders, 
which leads to degraded accuracy in sewing-pattern generation and high computational overhead. 
Second, most frameworks depend on commercial or proprietary physics engines~\cite{Qualoth, Style3D, CLO3D} or on open-source garment modeling frameworks such as GarmentCode~\cite{gc}, built on NVIDIA Warp~\cite{warp}, 
which are computationally expensive and slow. 
These simulators must iteratively stitch 2D panels under physical constraints and apply gravity and collision forces to drape garments on human models, 
making them inefficient for scalable generation.

Revisiting the garment creation process, we wonder: \textbf{can we synthesize high-quality and coherent 3D garments without any physics simulation in the sewing process?} 
Inspired by the structured nature of industrial garment design, we propose \textbf{SwiftTailor}, an efficient two-stage framework that follows the real-world garment production pipeline: first generating sewing patterns, and then constructing 3D garments. Our approach bypasses physics simulation in the sewing stage, while generated 3D garments remain compatible with downstream simulation tasks.

At the core of our framework is a novel compact garment representation called the Garment Geometry Image (GGI), 
which represents 3D garment meshes in a unified UV texture space. 
In the first stage, we propose PatternMaker, a lightweight multimodal large language model trained to predict sewing patterns from textual or visual descriptions, 
achieving superior accuracy compared to larger VLM baselines such as AIpparel~\cite{aipparel} and ChatGarment~\cite{chatgarment}. 
In the second stage, we propose GarmentSewer, an efficient dense prediction transformer network that converts the generated sewing patterns in the first stage into dense GGI representations that encode local 3D geometry for all garment panels, 
from which the final mesh can be reconstructed through an efficient inverse-mapping process with remeshing and dynamic stitching~\cite{sander2003multi}, 
eliminating the need for a physical re-simulation of sewing.

Our contributions are summarized as follows:
\begin{itemize}
    \item We introduce the Garment Geometry Image, a novel representation that transforms 2D sewing patterns into 3D garment meshes without any physics-based solver.
    \item We present a modular and efficient pipeline comprising PatternMaker and GarmentSewer, achieving faster sewing-pattern reasoning and real-time garment construction compared to existing frameworks.
    \item Our approach supports multiple input modalities (text or image) and multiple downstream tasks (generation, and editing), achieving state-of-the-art results on the GarmentCodeData benchmark~\cite{gcd} with significantly reduced computational cost.
    \item We demonstrate a unified pipeline in which each stage is designed to be modular and can be easily integrated into existing garment generation methods.
\end{itemize}    
\section{Related Works}

\noindent \textbf{Pattern Generation.}
Early studies on digital garment modeling focus on parametric representations, where each garment is described by a compact set of geometric parameters and explicit stitching relations. 
This formulation allows pattern generation through sampling in the parameter space and reconstructing 3D garments via predefined sewing rules \cite{gc, oldgcd, gcd}. 
Among them, GarmentCode~\cite{gc} provides a programmable interface that represents garments as structured sewing programs. 
By sampling in its parameter space, new patterns can be generated efficiently while maintaining structural validity. 
However, this parametric method remains limited when garments must satisfy multiple high-level conditions, such as visual appearance, textual description, or style preferences, which require a deeper semantic understanding of design intent.

To overcome these limitations, recent approaches employ large vision-language or autoregressive models to predict sewing patterns directly from multimodal inputs \cite{aipparel, chatgarment, design2garment, he2024dresscode, sewformer}. These models treat pattern elements including panels, edges, and stitching connections as tokens, and learn to reason over their layout and geometry through sequential generation. Such methods improve semantic controllability and support interactive tasks like text-guided design and editing, yet their large backbones or inefficient representations often lead to high inference cost and reduced geometric precision.

In parallel, diffusion-based models have emerged as an alternative for controllable pattern synthesis. They learn to generate sewing patterns by iteratively denoising a compact representation of the pattern, enabling diverse and fine-grained sampling under complex multimodal conditions~\cite{sewingldm, garmagenet, garmentdiffusion}, and also achieving faster inference than the autoregressive counterparts. However, these models are tightly coupled to a predefined representation with a fixed size, which becomes a limitation when the representation is later extended - such as adding support for accessories or new components - requiring models retrained from scratch.

\noindent \textbf{Garment Construction.}
The conversion from 2D sewing patterns to 3D garments has traditionally relied on physics-based solvers. Commercial systems~\cite{CLO3D,Style3D,Qualoth} achieve realistic draping, yet remain closed-source. Open-source alternatives, including GarmentCode~\cite{gc}, built on NVIDIA Warp~\cite{warp}, employ GPU-accelerated solvers based on XPBD~\cite{xpbd}, or C-IPC~\cite{cipc}, or more recent Newton framework~\cite{newton}.
While accurate, these iterative simulations are computationally intensive and difficult to scale for generative modeling. Moreover, both commercial and open-source pipelines often require manual adjustment to correctly align panels before sewing. Although GarmentCode~\cite{gc} provides heuristic initialization rules for panel registration, these rules are not generalizable and may result in distorted garments, reported in ~\cite{gcd}.
To address these limitations, recent learning-based approaches~\cite{garmagenet, garmentimage, neuralsewingmachine, tailornet, deepwrinkles, garment3dgen, wang2025garmentcrafter, li2025single, li2024garment} aim to directly construct coherent 3D garments from 2D patterns. These methods demonstrate the potential of integrating physical consistency into data-driven models, which also motivates our approach.    
\section{Preliminaries}
\subsection{Sewing Pattern}

A sewing pattern defines the 2D blueprint of a garment, consisting of a collection of planar panels and a set of stitching relationships. Each panel corresponds to a specific region of the 3D garment surface, with a known placement around the human body, and is annotated with edge information that determines potential boundary connections. Following \cite{gc, gcd, aipparel}, we represent a sewing pattern $\mathcal{P} = (\mathbf{P}, \mathbf{S})$ as
\begin{equation}
\begin{aligned}
\mathbf{P} &= \big\{P_i = (V_i, E_i, R_i) \big\}_{i=1}^{N}, \\
\mathbf{S} &= \big\{\, s_k = (e_a, e_b) 
      \mid e_a, e_b \in \cup_{i=1}^{N} E_i \,\big\}_{k=1}^{M},
\end{aligned}
\end{equation}
where $\mathbf{P}$ comprises $N$ panels $P_i$, each defined by a set of vertices $V_i$, edges $E_i$, and a rigid transformation $R_i$, which determines the placement of the panel in 3D space when sewing. 
Each panel forms a self-closed loop, such that the number of vertices equals the number of edges 
($|V_i| = |E_i|$). 
$\mathbf{S}$ denotes the global set of $M$ stitching pairs. 
Each stitching pair $s_k = (e_a, e_b)$ specifies two boundary edges, either from the same panel (e.g. darts or pleats) or from different panels (inter-panel seams) that must be merged during garment assembly. 
Applying all stitching operations in $\mathbf{S}$ reconstructs the complete 3D garment topology, 
ensuring structural continuity across panels.

Unlike conventional 3D mesh representations that directly store surface connectivity in Euclidean space, sewing patterns separate geometric shape from structural topology. 
This separation offers several advantages: 
(1) the panels are compact and well-structured in 2D, which can be directly used as UV domains for downstream tasks such as geometry or texture mapping, while also facilitating efficient data storage; 
and (2) the sewing patterns and their associated operations are explicitly defined and fully compatible with industrial garment-design workflows employed in real production environments.

\subsection{Geometry Image}

A Geometry Image (GIM)~\cite{gu2002geometry} represents a 3D surface in a 2D image-like format by defining a mapping function \( f: \mathcal{S} \rightarrow [0, 1]^2 \), where \( \mathcal{S} \subset \mathbb{R}^3 \) denotes the 3D surface and \( [0, 1]^2 \) the UV domain. Reconstructing the original surface requires the inverse mapping \( f^{-1} \), which recovers both 3D coordinates and mesh connectivity. Building an effective GIM involves two key steps: generating the image from the surface using \( f \), and recovering the mesh through remeshing with (an approximate) \( f^{-1} \).

For more complicated shapes, the Multi-chart Geometry Image (MCGIM)~\cite{sander2003multi} extends this concept by leveraging an atlas representation that partitions the surface into a geometrically natural set of charts. Each chart is individually parameterized onto an irregular polygon and then packed together into a single geometry image. MCGIM~\cite{sander2003multi} also introduces the zippering scheme as a part of inverse mapping  \( f^{-1} \) to reconnect charts.
Previous works, such as AtlasNet~\cite{atlasnet}, Omage~\cite{yan2024omages64}, or Geometry Image Diffusion~\cite{gimdiffusion}, have exploited this representation for 3D generation, but without defining an explicit zippering scheme, which results in cracks in final meshes - a problem also discussed in \cite{patchstitching}. More recent work for 3D garment generation, GarmageNet~\cite{garmagenet}, considers this issue by introducing an additional network that learns to stitch charts together. However, since the geometry image values in their framework do not accurately reflect the true 3D shape of the garment, they still require a further step using a simulation engine to resimulate the sewing process.
\begin{figure}[t]
    \centering
  \includegraphics[width=\linewidth]{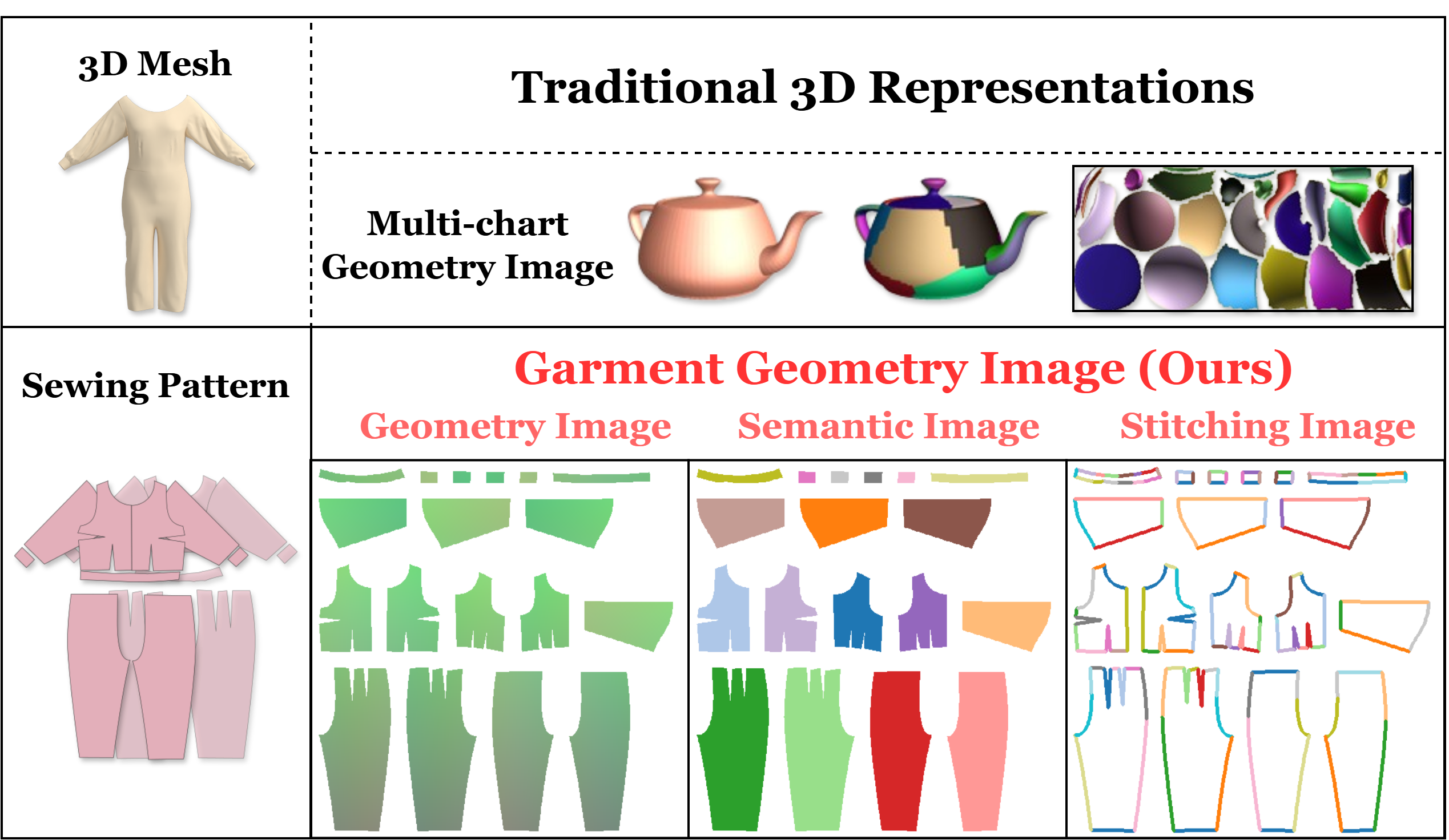}
  \caption{
    Preliminaries on geometry images~\cite{gu2002geometryimage,sander2003multi}, an image-based 3D representation that parameterizes a 3D mesh into charts, each being stored as simple arrays of pixels.  
    Our work integrates geometry images with semantic and stitching information to establish garment panels, yielding a novel garment geometry image representation suitable for 3D garment generation. 
  }
  \label{fig:preliminaries}

\end{figure}
\begin{figure*}[t]
    \centering
  \includegraphics[width=\textwidth]{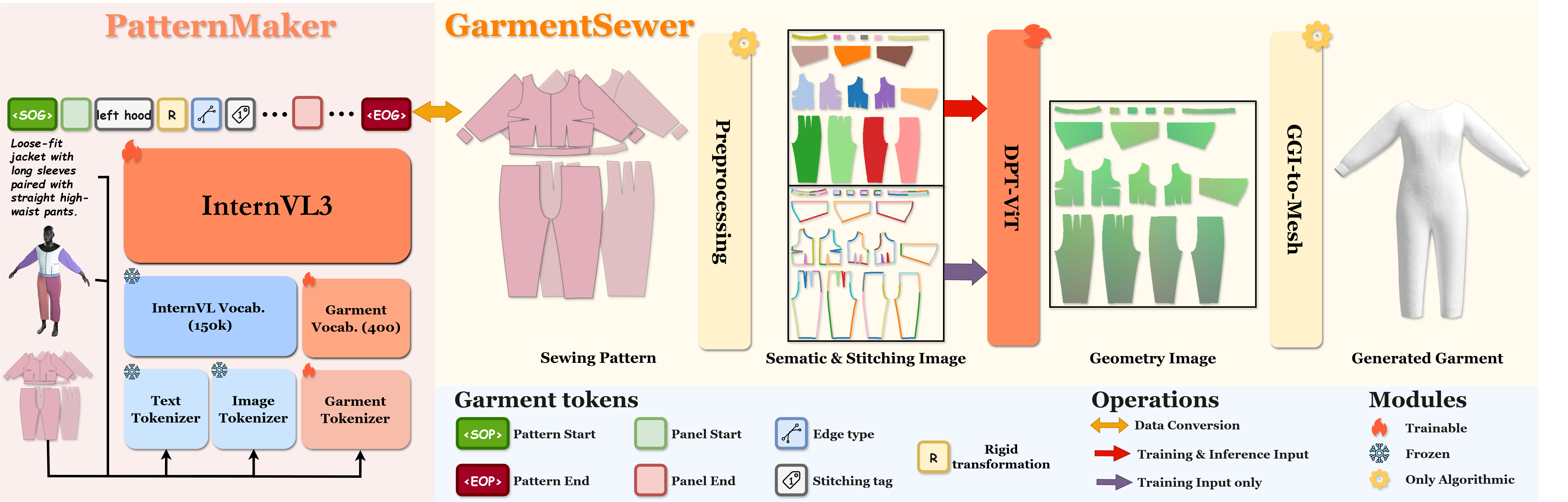}
  \caption{
    Overall pipeline. Our PatternMaker is a relatively small vision-language model (InternVL-3-2B~\cite{wang2025internvl3}) trained to output sewing patterns. 
    The sewing patterns are constructed from discrete tokens and continuous parameters predicted by the VLM. 
    Our GarmentSewer is a dense prediction transformer (DPT) that predicts a garment geometry image from the sewing patterns. In this step, we preprocess the sewing pattern to achieve the semantic and stitching map, which are then passed to the DPT to predict the geometry image, completing our garment geometry image representation (GGI). We then perform a postprocessing step to convert the GGI to a final 3D mesh. 
  }
  \label{fig:overall-pipeline}
  \vspace{-10pt}
\end{figure*}
\section{Methodology}

\noindent \textbf{Problem Statement.} 
Given reference images or a textual description of a garment, our objective is to generate its 3D mesh without relying on physics-based simulation software for garment assembly.

As can be seen in the Fig.~\ref{fig:overall-pipeline}, we decompose the garment construction process into two sequential stages. In Section~\ref{pattern_marker_section}, we first introduce \textbf{PatternMaker}, a lightweight multimodal large language model (MLLM) trained to generate the garment’s sewing pattern $\mathcal{P}$. Previous works~\cite{aipparel,chatgarment,sewingldm,design2garment, garmentimage} then utilize the GarmentCode~\cite{gc} engine to obtain the 3D mesh from $\mathcal{P}$. This procedure requires several sub-processes to lift the predicted panels to 3D and then sew them into a continuous mesh by simulating sewing forces and handling collision. We replace this stage by introducing a feed-forward neural network, \textbf{GarmentSewer}, to directly obtain the 3D mesh (see Section~\ref{garment_sewer_section}).
To bridge the substantial gap between the two distinct representations (sewing patterns $\mathcal{P}$ and 3D meshes), we further propose the \textbf{Garment Geometry Image (GGI)} (see Section~\ref{ggi_section}), an intermediate yet essential representation that enables accurate and efficient conversion from sewing patterns to 3D meshes.

\subsection{PatternMaker}\label{pattern_marker_section}
Given multimodal inputs (images or text), PatternMaker generates the sewing pattern~$\mathcal{P}$. We adopt the sewing-pattern representation from GarmentCode~\cite{gcd}, which encodes discrete structure (panel layout, edge connectivity, stitching tags) and continuous geometry (vertex coordinates and rigid transformations). PatternMaker itself is agnostic to the specific tokenization scheme, but we use AIpparel’s representation for its simplicity and expressive power~\cite{aipparel}.

While AIpparel~\cite{aipparel} and Chat Garment~\cite{chatgarment} fine-tune the 7B-parameter LLaVA-1.5 model~\cite{liu2023llava}, we instead train the much smaller InternVL-3-2B model~\cite{wang2025internvl3}. We retain the tokenizer and MLP regression heads from AIpparel for a fair comparison. Despite using only 30\% of the parameters, PatternMaker achieves significantly higher pattern accuracy and topology validity. Further analysis and benchmarks are provided in Section~\ref{sew_results}. 

We also adopt the training losses of AIpparel~\cite{aipparel}. We jointly train discrete token prediction and continuous parameter regression. The newly introduced tokens for pattern generation are supervised by the next token prediction task, while the continuous parameters, including vertex positions and panel transformations, are predicted through small MLP regression heads.

 \begin{figure*}[t]
    \centering
  \includegraphics[width=\textwidth]{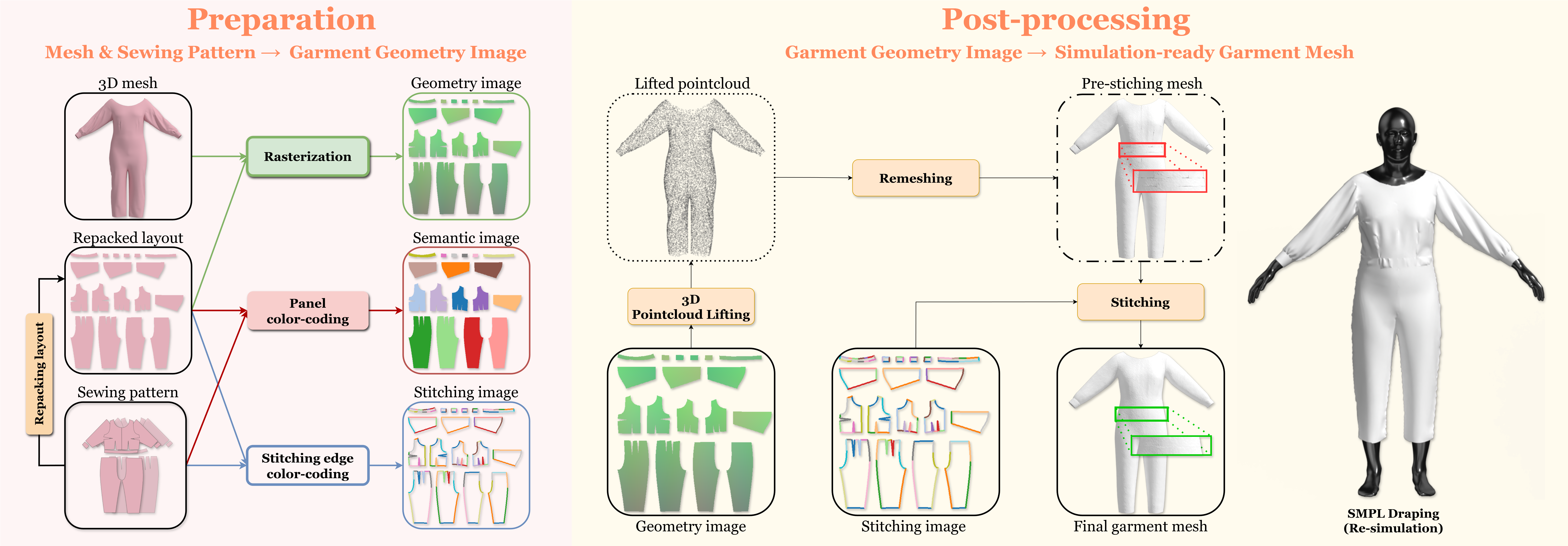}
  \caption{ 
    (\textit{Left}) We present how to prepare the three components (geometry, semantic and stiching) of our propose Garment Geometry Image (GGI); (\textit{Right}) From the estimated geometry and stiching images of GarmentSewer and PatternMaker, two additional remeshing and stiching steps are performed to obtain the final 3D mesh result. 
  }
  \label{fig:forward-backward-pipeline}
  \vspace{-10pt}

\end{figure*}

\subsection{Garment Geometry Image} \label{ggi_section}

\label{subsec:garment-geometry-image}
In the following section, we bridge gap between the sewing pattern from PatternMaker and the 3D garment mesh produced by GarmentSewer. These representations are fundamentally different: the pattern is a structured, discrete description (panels, vertices, stitching relations) in a serialized format~\cite{gc,gcd,oldgcd}, whereas the 3D mesh is a dense, continuous surface defined by point clouds and faces, making direct conversion difficult and unstable.

\noindent \textbf{Connections with Geometry Images}
Recent works~\cite{gimdiffusion,yan2024omages64,atlasnet} show that MCGIM offers an image-like representation that enables efficient 3D learning with standard 2D architectures and straightforward mesh reconstruction. Sewing patterns, however, provide richer semantics (e.g., torso, cuff) and explicit stitching relations that are ignored in those approaches. Motivated by these complementary advantages, we propose the Garment Geometry Image, a unified representation that combines the geometric structure of MCGIM with the semantic and stitching priors of sewing patterns.

\noindent \textbf{Definition.}
The Garment Geometry Image consists of three aligned components, comprising a \textit{semantic image}, a \textit{geometry image}, and a \textit{stitching image}, which share a common repacked panel layout from the sewing pattern (see~\cref{fig:preliminaries}). The semantic image encodes panel types via color, the geometry image stores 3D surface coordinates in pixel space, and the stitching image records edge-wise stitching information, where edges with the same color are sewn together in post-processing. This unified representation tightly links 2D pattern reasoning to 3D mesh construction, effectively connecting PatternMaker and GarmentSewer.

\subsection{GarmentSewer}\label{garment_sewer_section}

In the following section, we first describe the process for generating geometry, semantic and stitching images, which are essential components of the GarmentSewer pipeline. We then delve into the model architecture, followed by a discussion of the training scheme used to achieve high-quality 3D garment mesh generation.

\noindent \textbf{GGI Preparation.}
As shown in Fig.~\ref{fig:forward-backward-pipeline} (left), we first preprocess the estimated sewing patterns to obtain \textit{the semantic and stitching images}. We perform a repacking step that arranges all garment panels into a densely packed square layout. Using the sewing-pattern metadata, we then color-code panel types and boundary edges to generate the semantic and stitching images, respectively.

Inspired by seminal works~\cite{pix2pix2017,saharia2022palette, dpt} on image-to-image translation, we design GarmentSewer as a mapping function from the semantic image to the geometry image. The geometry image serves as an intermediate representation from which we construct the 3D garment mesh in a subsequent post-processing step. For the \textit{geometry image}, we normalize 3D vertex coordinates and rasterize each garment-mesh vertex to its corresponding pixel location via UV mapping. However, the number of mesh vertices is much smaller than the number of pixels in the geometry image, leading to sparsely sampled artifacts. To address this, we apply a hybrid interpolation strategy that combines linear and barycentric interpolation to fill missing pixel values and produce a smooth, continuous geometry image.

\noindent \textbf{Model Training} GarmentSewer follows the standard DPT~\cite{dpt} architecture, consisting of a ViT-based encoder~\cite{khan2022transformers} and a multi-scale convolutional decoder. The encoder extracts both global garment layout and fine-grained panel structure from the semantic image, while the decoder fuses hierarchical features to reconstruct dense geometry at multiple scales. We train the model using three losses:

\begin{table*}[!h]
\centering
\caption{Quantitative results on sewing-pattern generation (left) and editing (right). Best results are shown in \textbf{bold}.}
\label{tab:pattern_prediction}
\resizebox{\linewidth}{!}{
\begin{tabular}{lcccccc}
\toprule
Method & Vertex L2 ($\downarrow$) & \#Panel Acc ($\uparrow$) & \#Edge Acc ($\uparrow$) & Rot L2 ($\downarrow$) & Transl L2 ($\downarrow$) & Stitch Acc ($\uparrow$) \\
\midrule
AIpparel~\cite{aipparel} & 4.8 / 2.5 & 93.7 / 82.9 & 79.0 / 88.2 & 0.007 / 0.002 & 2.5 / 1.7 & 73.0 / 86.3 \\
ChatGarment~\cite{chatgarment} & 14.9 / 13.5 & 16.4 / 19.6 & 49.7 / 60.3 & 0.038 / 0.036 & 15.4 / 10.1 & 38.2 / 55.2 \\
SewingLDM~\cite{sewingldm} & 15.6 & 18.0 & 49.0 & 0.052 & 16.6 & 30.6 \\
\midrule
\textbf{PatternMaker (Ours)} & \textbf{3.5} / \textbf{1.5} & \textbf{94.8} / \textbf{85.0}  & \textbf{92.3} / \textbf{98.0} & \textbf{0.006} / \textbf{0.002} & \textbf{1.9} / \textbf{1.3}  & \textbf{85.1} / \textbf{97.8} \\
\bottomrule
\end{tabular}}
  \vspace{-10pt}

\end{table*}
\begin{itemize}

    \item Regression loss: Geometry images are inherently edge-sensitive, so pixels near panel boundaries are given higher weights. Given ground-truth geometry image $\mathcal{G}$ and prediction $\hat{\mathcal{G}}$, the edge-aware L1 regression loss is
\begin{equation}
    \mathcal{L}_{\text{reg}}
    = 
    \underbrace{\| \mathcal{G} - \hat{\mathcal{G}} \|_{1}}_{\text{interior supervision}}
    + 
    \underbrace{
    \alpha \left\| \mathcal{G}_{\text{edge}} - \hat{\mathcal{G}}_{\text{edge}} \right\|_{1}
    }_{\text{edge proximity band (width $w$= 10)}}
    \label{eq:edge-aware-regression}
\end{equation}
where $\mathcal{G}_{\text{edge}}$ and $\hat{\mathcal{G}}_{\text{edge}}$ denote pixels extracted from the union of all edge-proximity bands:
$\mathcal{G}_{\text{edge}} = \bigcup_{i=1}^{n} \mathrm{prox}(E_i, w)$. Here $\alpha$ is a weighting factor balancing interior and edge-sensitive supervision.
    \item Stitching loss: To enable garment assembly without re-simulating sewing, stitched panel edges must closely align in 3D. Using the stitching map to identify paired edges $(e_a, e_b)$, we compute a Chamfer-distance (CD) loss between their predicted boundary points:
\begin{equation}
    \mathcal{L}_{\text{stitch}}
    =
    \frac{1}{|\mathbf{S}|}
    \sum_{(e_a, e_b) \in \mathbf{S}}
    \mathrm{CD}\big(
        \hat{\mathcal{G}}_{\text{edge}}(e_a),
        \hat{\mathcal{G}}_{\text{edge}}(e_b)
    \big)
    \label{eq:stitch}
\end{equation}
    \item Finally, we adopt the normal-regularization term from~\cite{turkulainen2025dn} to ensure that the surface of the generated 3D garment mesh remains smooth.

\end{itemize}

\noindent \textbf{Postprocessing.}
Here, we outline the steps for converting the predicted geometry image (from GarmentSewer) and stitching image (from PatternMaker) into a 3D mesh (see Fig.~\ref{fig:forward-backward-pipeline}, right). We first lift the geometry image to 3D to obtain a garment point cloud. A remeshing step then reconstructs individual panel surfaces, and a stitching step restores global connectivity, producing a unified 3D garment mesh suitable for simulation or rendering. Further details on remeshing and stitching are provided in the appendix.

\section{Experiments}

\subsection{Experimental Settings}
\noindent \textbf{Trainng Details.} 
PatternMaker is obtained by fine-tuning InternVL-3-2B~\cite{wang2025internvl3} for sewing-pattern generation, following the tokenizer and regression heads of AIpparel~\cite{aipparel}. 
GarmentSewer uses a DPT architecture~\cite{dpt} with a ViT-L encoder~\cite{vit} initialized from ImageNet~\cite{imagenet}. 
We train it with the loss terms using 
$\lambda_{\text{reg}}{=}1$, 
$\lambda_{\text{stitch}}{=}1000$, 
$\lambda_{\text{norm}}{=}0.01$, 
and $\alpha{=}100$. Each model is trained on 4 A100 GPUs within 3 days. 
Further hyperparameters and implementation details are provided in the Supplementary.

\noindent \textbf{Datasets.} We train our models on GCD-MM~\cite{aipparel}, the multimodal extension of GarmentCodeData~\cite{gcd}. We follow the same train-validation-test split as defined in GCD-MM. For SewingLDM~\cite{sewingldm}, we follow their instruction to extract the garment sketch as input for the model.

\subsection{Sewing Pattern Generation} \label{sew_results}
\noindent \textbf{Task setup.}
Given multimodal inputs, this stage’s goal is to generate pattern panels 
$\mathbf{P}$ and their associated stitching set 
$\mathbf{S}$. The resulting representation defines both the geometric layout of each 2D panel and the stitching relationships required for 3D reconstruction.

\noindent \textbf{Baselines.} 
We compare PatternMaker with recent multimodal models for sewing-pattern reasoning, including 
AIpparel~\cite{aipparel}, ChatGarment~\cite{chatgarment}, and SewingLDM~\cite{sewingldm} for multi-modal generation, while editing task is done on the first two models, as SewingLDM does not support this task.

\noindent \textbf{Evaluation metrics.}
Following prior works \cite{aipparel, sewingldm}, we evaluate the accuracy of both the discrete structure (accuracy on \#Edges, \#Panels, and Stitching) and the continuous parameters (including rigid transformations with Rotation L2, and Translation L2, and Vertex L2 for coordinates ).

\noindent\textbf{Results.} 
As shown in \cref{tab:pattern_prediction}, PatternMaker surpasses all baselines, achieving lower geometric error and higher stitching accuracy than AIpparel~\cite{aipparel} despite using a smaller backbone.
This improvement comes from fine-tuning the more efficient InternVL-3-2B~\cite{wang2025internvl3} model in place of the heavier LLaVA-1.5V-7B~\cite{liu2023llava}. 
ChatGarment~\cite{chatgarment} and SewingLDM~\cite{sewingldm} underperform due to weaker structural reasoning. 
These results highlight that an efficiently fine-tuned multimodal model can outperform larger counterparts while remaining computationally lightweight.

\noindent\textbf{Sewing Pattern Editing.} For the text-guided editing task~\cite{aipparel}, according to \cref{tab:pattern_prediction},  PatternMaker also achieves the highest structural and transformation accuracy.
It reliably interprets fine-grained instructions while maintaining valid panel geometry and seam topology.

\subsection{Garment Mesh Generation}

\begin{figure*}[t]
    \centering
  \includegraphics[width=0.9\linewidth]{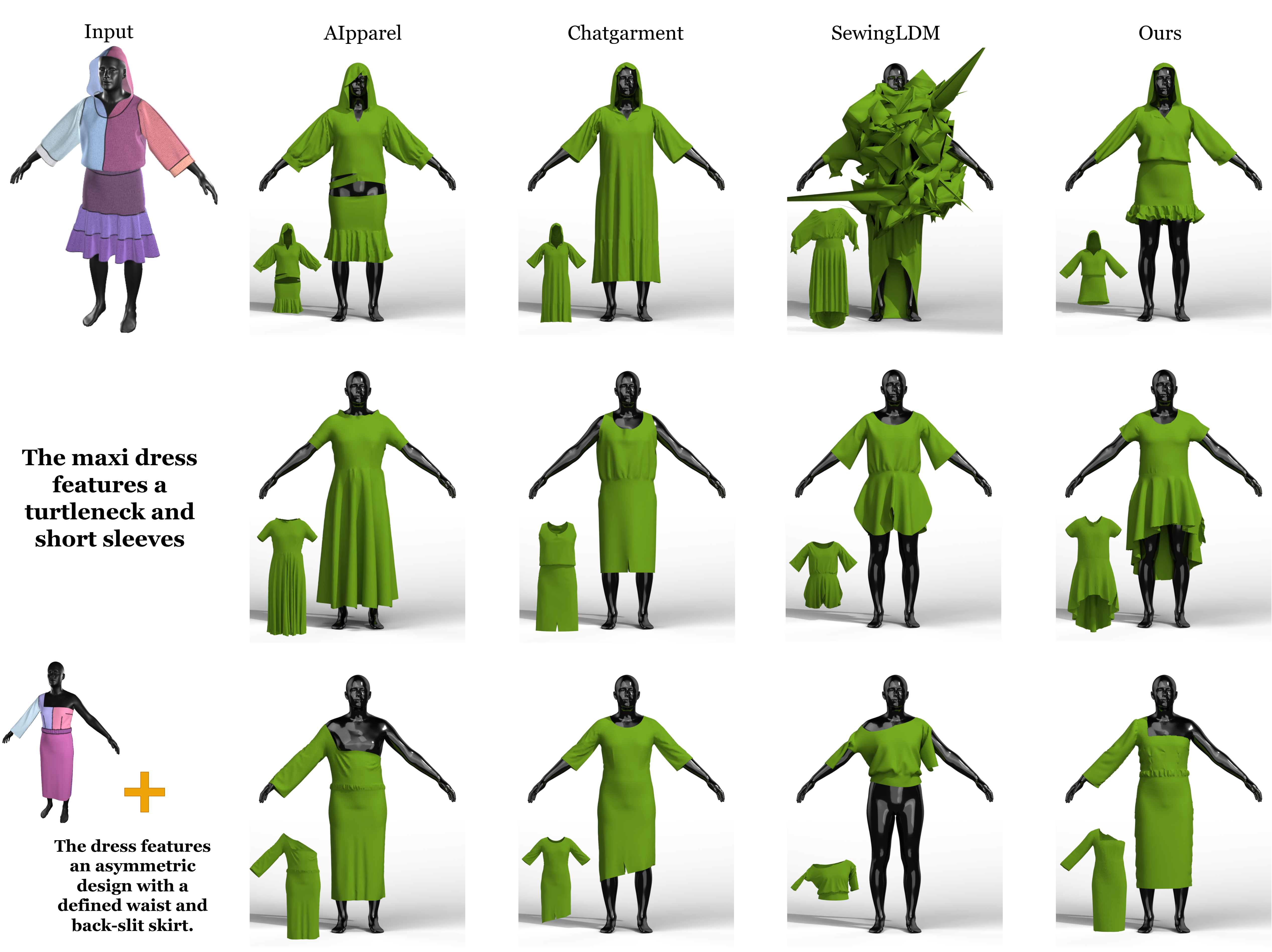}
  \caption{
    Qualitative comparisons between SwiftTailor and recent state-of-the-art methods on 3D garment modeling~\cite{aipparel,chatgarment,sewingldm} using an image, a text prompt, and both text and image as input, respectively.
  }
  \label{fig:qualitative}\
  \vspace{-10pt}
\end{figure*}

\noindent \textbf{Task setup.}
Given a sewing pattern, this stage evaluates the model's ability to construct a 3D garment mesh. We assess performance along two dimensions: (1) the quality of generated meshes relative to reference meshes, and (2) the computational cost of converting sewing patterns into 3D meshes. We also provide qualitative comparisons between our SwiftTailor pipeline and alternative methods paired with GarmentCode (\cref{fig:qualitative}), showing that GarmentSewer produces more reliable initial states before simulation, avoiding bending or failure. All metrics in this section are computed directly from \textbf{GarmentSewer’s output}, not from meshes obtained after physical simulation.

\noindent\textbf{Baselines.} We benchmark the same set of methods from the previous section, using the GarmentCode engine for garment construction, and compare them against our full SwiftTailor pipeline, which replaces GarmentCode with GarmentSewer as the construction module.

\noindent\textbf{Evaluation Metrics.} 
Following prior works in diffusion-based point cloud generation~\cite{pcdiff} and garment generation~\cite{garmagenet}, we evaluate garment generation quality using Minimum Matching Distance (MMD) and Coverage (COV). We compute Chamfer Distance between point clouds for distance-based metrics. Besides the metrics on 3D generation, we also report average number of sampling as computational cost to get the first successful conversion from sewing pattern into 3D mesh. For the generation and construction time, we report in a separate table.

\noindent\textbf{Garment Generation.} The quantitative evaluation in \cref{tab:multi_condition} shows that SwiftTailor achieves the best MMD and highest COV, indicating that it produces both higher-quality and more diverse garments. Our GarmentSewer module also outperforms physics-based construction via GarmentCode~\cite{gc}, and integrating PatternMaker with GarmentSewer yields substantially stronger results than pairing PatternMaker with GarmentCode~\cite{gc}. Moreover, SwiftTailor is highly robust: it requires fewer sampling attempts to obtain a valid 3D garment, second only to ChatGarment. While ChatGarment benefits from predicting coarse, high-level garment attributes and then relying on GarmentCode for pattern sampling~\cite{chatgarment}, this coarse formulation limits its fine-grained control and accuracy, leading to weaker generation quality compared to our more explicit pattern specification and geometry-aware pipeline.
\begin{table}[t]
\centering
\small
\caption{Quantitative results on mesh generation using multi-modal inputs (image and text). Best results are shown in \textbf{bold}.}
\label{tab:multi_condition}
\resizebox{\linewidth}{!}{
\begin{tabular}{lcccc}
\toprule
Method & MMD$\downarrow$ & COV$\uparrow$ & \#Sampling$\downarrow$ \\
\midrule
AIpparel~\cite{aipparel} + GC~\cite{gc} & 6.94 & 0.52 &  4.27 \\
ChatGarment~\cite{chatgarment} + GC~\cite{gc} & 12.27 &	0.22 &  \textbf{1.20} \\
SewingLDM~\cite{sewingldm} + GC~\cite{gc} & 11.33 & 0.34	&  5.87 \\
\midrule
\textbf{PatternMaker + GC~\cite{gc}} & 6.82 & 0.54 & 2.98 \\
\textbf{SwiftTailor (Ours)} & \textbf{5.31} & \textbf{0.68} & 2.98 \\
\bottomrule
\end{tabular}
}

\end{table}

\begin{table}[t]
\centering
\small
\caption{Running time comparison  to obtain the final mesh (in seconds) between other baselines and our SwiftTailor. Stage 1 is generating patterns, while Stage 2 is constructing mesh from them.
}

\label{tab:running_time}
\resizebox{\linewidth}{!}{
\begin{tabular}{lcccc}
\toprule
Method & Stage 1 & Stage 2 & Post-proc. & Total \\
\midrule
AIpparel~\cite{aipparel} + GC~\cite{gc} & 10.20 & 49.55 & 3.99 & 63.74\\
ChatGarment~\cite{chatgarment} + GC~\cite{gc} & 18.85 & 33.92 & 5.76 & 58.53 \\
SewingLDM~\cite{sewingldm} + GC~\cite{gc} & \textbf{5.44} & 46.87 & \textbf{2.16} & 54.47 \\
\midrule
\textbf{PatternMaker + GC~\cite{gc}} & 9.93 & 37.45 & 2.18 & 49.56 \\
\textbf{SwiftTailor (Ours)} & 9.93 & \textbf{0.02} & $\text{4.83}$ & \textbf{14.78}\\
\bottomrule
\end{tabular}
}
\end{table}
\noindent\textbf{Running time.} Without relying on physical simulation (e.g., GarmentCode) for mesh generation, our method achieves a significant speedup, as shown in \cref{tab:running_time}. Specifically, our Stage 2 achieves an inference time of 0.02s, which is orders of magnitude faster than all baselines using GarmentCode~\cite{gc}. Consequently, considering the total inference time across all stages, our pipeline generates simulatable garments roughly 4× faster, demonstrating that it can produce high-fidelity garments with higher efficiency.
\noindent\textbf{Qualitative examples.}
\cref{fig:qualitative} presents qualitative comparisons across all input modalities. 
For image-conditioned inputs, our method best matches the reference images, 
while AIpparel~\cite{aipparel} produces meshes with missing seams causing tearing artifacts, 
and SewingLDM~\cite{sewingldm} generates degenerated meshes that fail simulation. 
For text-conditioned inputs, our results are on par with existing baselines. 
For combined text-image inputs, although AIpparel~\cite{aipparel} and SewingLDM~\cite{sewingldm} can output 3D meshes, their draping on SMPL~\cite{smpl} is often misaligned 
due to the rule-based physical simulation in GarmentCode~\cite{gc}, reducing reliability for downstream use. 
Across modalities, ChatGarment~\cite{chatgarment} produces drapable garments but with limited style diversity and condition mismatches. 
In contrast, our approach consistently yields high-quality, stable, and diverse garments, confirming the effectiveness of our design.

\subsection{Ablation Studies}
In the ablation studies, we focus on our main contributions: the GGI representation and the training strategy for GarmentSewer. We first demonstrate the necessity of the semantic image as conditioning input for GarmentSewer to generate the geometry image. We then ablate two key losses, namely regression and stitching, to evaluate their individual impact on the final results. Each ablation is analyzed both qualitatively and quantitatively.

\begin{figure}[t]
    \centering
  \includegraphics[width=\linewidth]{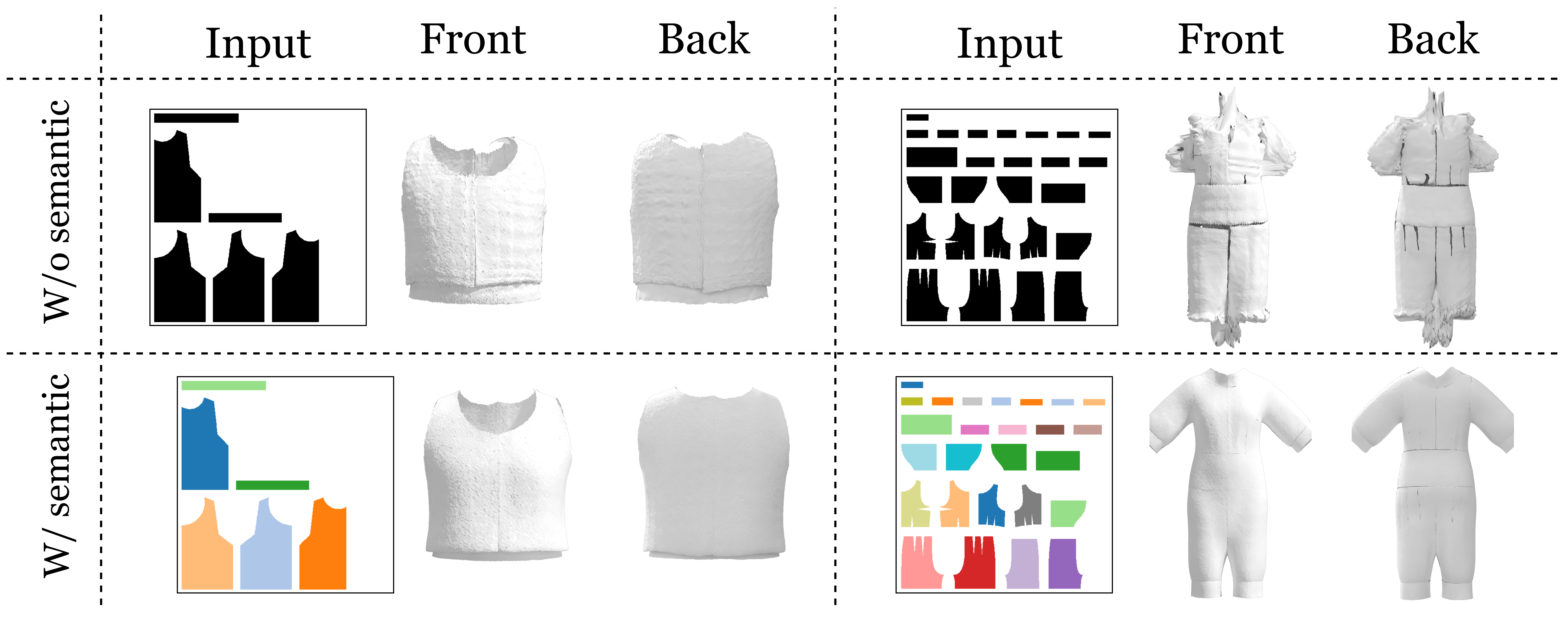}
  \caption{
    Qualitative results on generated 3D mesh using geometry vs. binary image as input to GarmentSewer.    
}
  \label{fig:ablation-semantic-map}
  
\end{figure}

\begin{table}[t]
    \centering
    \setlength{\tabcolsep}{6pt}
        \caption{Ablation on semantic UV map and  auxiliary losses}
    \resizebox{\linewidth}{!}{
    \begin{tabular}{lcccc}
        \toprule
        Ablation & CD$\downarrow$ & EMD$\downarrow$ & MMD$\downarrow$ & COV$\uparrow$ \\
        \midrule
        \textit{W/o Semantic UV Map} &  35.77 & 9.90 & 11.96 & 0.49 \\
        \textit{W/ Semantic UV Map} & &  & \\
        \quad + $\mathcal{L}_{\text{reg}}$ &  9.84 & 5.95 & 7.38 & 0.58 \\
        \quad + $\mathcal{L}_{\text{stitch}}$ (Ours) &  \textbf{3.40} & \textbf{4.48} & \textbf{3.36} & \textbf{0.88}  \\
        \bottomrule
    \end{tabular}}
    \label{tab:uv_ablation}
    
\end{table}
\noindent\textbf{Semantic Image.} In this ablation, we examine the effectiveness of semantic image in training GarmentSewer. Using the same sewing-pattern layout, we train a variant that receives only a binary mask without panel-type encoding. 
While this version can handle simple cases with few panels, it fails to maintain correct topology in more complex garments (see~\cref{fig:ablation-semantic-map}). Without semantic cues, the model cannot reliably distinguish panels with similar shapes, such as left vs. right torso pieces or hood components, and struggles even more in multi-flare skirts where positional cues are minimal. 
In contrast, encoding panel types in the semantic map provides strong structural guidance, enabling accurate panel placement 
and yielding significantly better reconstruction performance (see~\cref{tab:uv_ablation}).

\begin{figure}[t]
  \label{fig:ablation-stage2-loss}
    \centering
  \includegraphics[width=\linewidth]{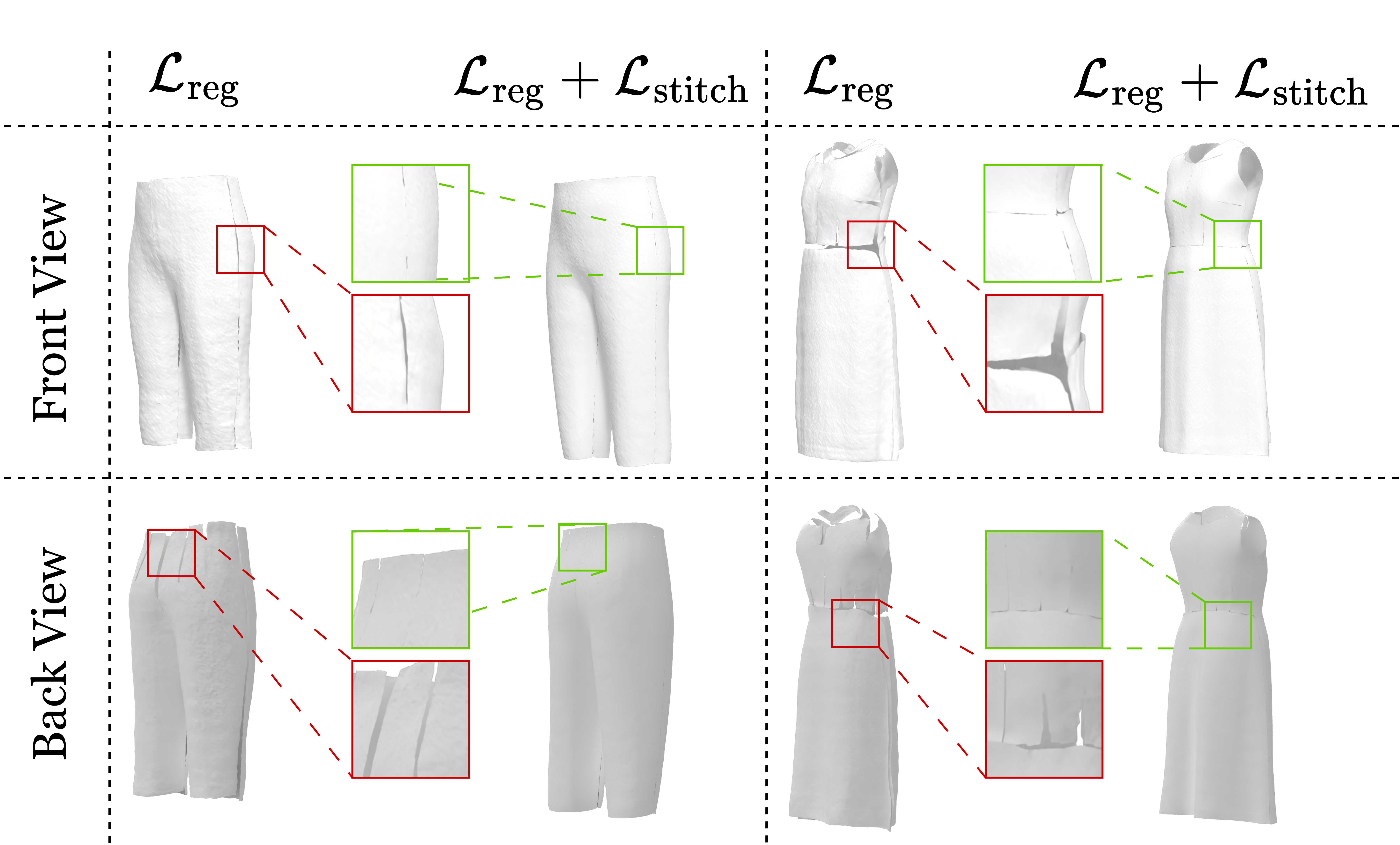}
  \caption{
    Qualitative results on ablation study of training losses.
  }
  
\end{figure}

\noindent\textbf{Garment Construction Losses.} As shown quantitatively in \cref{tab:uv_ablation} and qualitatively in \cref{fig:ablation-stage2-loss}, the edge-aware regression loss $\mathcal{L}_{\text{reg}}$ alone enables GarmentSewer to recover most of the garment shape, producing reasonably coherent panel geometry. 
However, without enforcing boundary consistency, the reconstructed meshes still exhibit noticeable seam gaps and misaligned edges. 
Introducing the stitching loss $\mathcal{L}_{\text{stitch}}$ effectively resolves these issues by aligning stitched panel boundaries, leading to substantial improvements across all metrics and visibly cleaner mesh connections. 
The normal loss $\mathcal{L}_{\text{normal}}$ is applied in all settings purely as a smoothness regularizer, but not driving the primary accuracy gains.

\section{Conclusion}
In this paper, we introduce \textbf{SwiftTailor}, a two-stage pipeline composed of \textbf{PatternMaker} and \textbf{GarmentSewer}. PatternMaker is a lightweight multimodal VLM that takes image or text inputs and predicts the garment’s sewing pattern. From this pattern, we introduce the \textbf{Garment Geometry Image (GGI)} as an intermediate representation that allows \textbf{GarmentSewer} to efficiently construct a simulation-ready 3D garment. This design improves both construction quality and computational efficiency over existing baselines. For future work, we plan to explore more efficient pattern-generation strategies for near real-time inference, and to investigate texture generation and wrinkle refinement without relying on physical simulation.

{
    \small
    \bibliographystyle{ieeenat_fullname}
    \bibliography{main}
}

\clearpage


\onecolumn
\appendix
\appendixpage
\counterwithin{figure}{section}
\counterwithin{table}{section}
\renewcommand{\contentsname}{\Large\textbf{Contents}}
\setcounter{tocdepth}{3}
\vspace{1em}
\begin{center}
\rule{0.8\textwidth}{1pt}
\end{center}
\tableofcontents
\vspace{1em}
\begin{center}
\rule{0.8\textwidth}{1pt}
\end{center}
\vspace{2em}

\clearpage

\section{Appendices Overview}

This appendix provides supplementary materials that support and extend the main paper. It is organized into three sections. In \cref{sec:ggi}, we describe the full procedure for preparing the Garment Geometry Image (GGI) along with the post-processing pipeline that converts a predicted or processed GGI back into a simulation-ready 3D garment mesh. In \cref{sec:experiment}, we provide detailed experimental settings, including dataset preparation, evaluation protocols, and additional qualitative and quantitative results to ensure fair comparison across baselines. In \cref{sec:discussions}, we discuss limitations, broader impact, and potential future directions for garment generation and 3D modeling. Together, these sections offer the technical details needed to reproduce our work and understand the complete scope of the SwiftTailor framework.

\section{Garment Geometry Image Preparation \& Postprocessing Pipeline}
\label{sec:ggi}

\subsection{Garment Geometry Image Preparation}

Following the overview in Sec.~4.3 of the main paper, we provide the complete procedure for constructing the Garment Geometry Image (GGI). The GGI is formed by three aligned components—geometry, semantic, and stitching images. Before generating these components, we first repack all garment panels into a unified square layout. This packed layout serves as the shared UV template onto which all information is embedded, ensuring alignment across components and enabling downstream tasks such as texture editing.

\begin{algorithm}[h]
\caption{Layout Packing with Orientation Correction}
\label{algo:repacking}
\begin{algorithmic}[1]
\Require Panels $\mathbf{P}$ in the sewing pattern
\Ensure Packed UV layout $L_{\text{UV}}$
\State $B \gets$ bounding-box sizes of all panels in $\mathbf{P}$
\State $B_{\text{sorted}} \gets$ sort $B$ by decreasing height, then width
\State Initialize binary-search range $[l, r]$ for the target square size
\While{$l < r$}
    \State $mid \gets \lfloor (l + r)/2 \rfloor$
    \State Test row-wise placement of $B_{\text{sorted}}$ inside a square of size $mid$
    \If{all panels fit}
        \State $r \gets mid$
    \Else
        \State $l \gets mid + 1$
    \EndIf
\EndWhile
\State $L_{\text{UV}} \gets$ final packing of $B_{\text{sorted}}$ within a square of size $l$
\For{each panel $P_i$ with bounding box $b_i$}
    \State Compute the outward-facing normal of $P_i$
    \If{the panel normal is opposite to the layout normal}
        \State Flip $P_i$ horizontally within $b_i$ and update $L_{\text{UV}}$
    \EndIf
\EndFor
\State \Return $L_{\text{UV}}$
\end{algorithmic}
\end{algorithm}

\paragraph{Layout Packing}
To obtain a compact and consistent representation, all panels predicted by the model are arranged within a single square layout. Given a set of panels~$\mathbf{P}$, we compute the bounding box of each panel and reduce the layout task to packing rectangles into a square of unknown size. A useful monotonic property holds: if all panels fit into a square of side length~$s$, then they also fit into any larger square~$s' > s$. This observation allows us to apply binary search to identify the smallest feasible square size, preventing unnecessary blank space that may hinder learning efficiency and waste memory storage. For each candidate size, we test feasibility using a simple row-wise packing heuristic: panels are sorted by height and width, and placed from bottom to top and from left to right. Although simple, this strategy is effective and fast across the GarmentCodeData~\cite{gcd}. The full algorithm is given in ~\cref{algo:repacking}.

A second consideration arises from panel orientation. Flattening all panels into UV space introduces inconsistencies in their outward-facing normals, especially between front- and back-facing panels. To enforce a consistent orientation across the layout, we choose a fixed normal direction for the whole layout and horizontally flip panels if mismatching. This ensures uniform facing direction in the packed UV map and simplifies later steps in remeshing.

\begin{algorithm}[h]
\caption{Semantic Image Creation from Packed Layout}
\label{algo:semantic-image}
\begin{algorithmic}[1]
\Require Panels $\mathbf{P}$ with type labels, packed layout $L_{\text{UV}}$, predefined color map $\mathcal{C}$ from panel types to unique colors
\Ensure Semantic image $GGI_{\text{semantic}}$
\State $GGI_{\text{semantic}} \gets$ blank image with the same resolution as $L_{\text{UV}}$
\For{each panel $P_i \in \mathbf{P}$}
    \State $R_i \gets$ UV region of $P_i$ in $L_{\text{UV}}$
    \State $t_i \gets$ type label of $P_i$
    \State $c_i \gets \mathcal{C}(t_i)$
    \State Fill region $R_i$ in $GGI_{\text{semantic}}$ with $c_i$
\EndFor
\State \Return $GGI_{\text{semantic}}$
\end{algorithmic}
\end{algorithm}

\paragraph{Semantic Image} After determining the packed UV layout, we generate the semantic image by assigning each panel a unique color based on its panel type, see~\cref{algo:semantic-image}. Using the metadata from the sewing pattern, every panel region in the packed layout is filled with its corresponding color, producing a dense map that encodes panel identity and functional category. This semantic image provides strong structural cues for GarmentSewer, enabling the model to distinguish between panels of similar shapes, such as left and right sleeves or front and back torso pieces, and to preserve correct topology in garments with other components.

\begin{algorithm}[h]
\caption{Stitching Image Creation from Packed Layout}
\label{algo:stitching-image}
\begin{algorithmic}[1]
\Require Panels $\mathbf{P}$ with stitched edge pairs $\mathbf{S}$, packed layout $L_{\text{UV}}$, predefined stitch color map $\mathcal{C}_{\text{stitch}}$
\Ensure Stitching image $GGI_{\text{stitch}}$
\State $GGI_{\text{stitch}} \gets$ blank image with the same resolution as $L_{\text{UV}}$
\For{each stitched pair $(e_a, e_b) \in \mathbf{S}$ with panels $(P_i, P_j)$}
    \State $B_a \gets$ boundary pixels of edge $e_a$ in $L_{\text{UV}}$
    \State $B_b \gets$ boundary pixels of edge $e_b$ in $L_{\text{UV}}$
    \State $k \gets$ stitch identifier of pair $(e_a, e_b)$
    \State $c_k \gets \mathcal{C}_{\text{stitch}}(k)$
    \State Color $B_a$ and $B_b$ in $GGI_{\text{stitch}}$ with $c_k$
\EndFor
\State \Return $GGI_{\text{stitch}}$
\end{algorithmic}
\end{algorithm}

\paragraph{Stitching Image}
Simultaneously, we construct the stitching image by encoding all boundary edges that participate in stitching relations. The boundary of each panel is extracted by applying a dilation operation on the packed layout to obtain a one-pixel-wide contour along its edges. For every stitched edge pair, we identify the corresponding boundary pixels and assign a shared stitch-identifier color to both edges, see~\cref{algo:stitching-image}. This produces a map in which all edges that must be joined in the final garment share the same color, enabling consistent boundary alignment and merge operations during postprocessing.

\begin{figure}[ht]
    \centering
  \includegraphics[width=\textwidth]{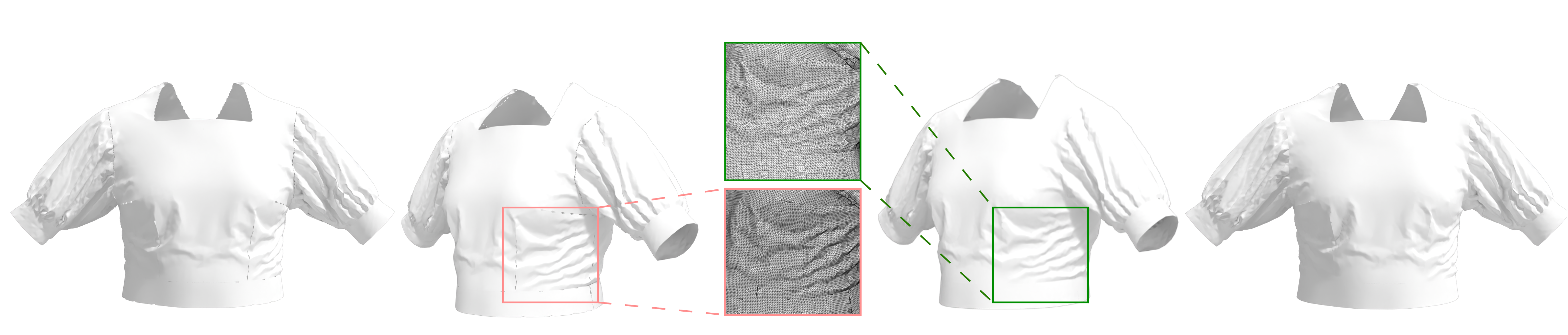}
  \caption{
    Effect of interpolation schemes on the geometry image. We first rasterize the mesh into a geometry image using different interpolation strategies and then remesh it back into 3D. Barycentric interpolation alone (\textit{left}) introduces jagged and discontinuous boundary values that deviate from the true panel contour. Our hybrid interpolation (\textit{right}), which applies linear interpolation along panel edges and barycentric interpolation only inside triangle interiors, produces smooth and consistent boundary signals, preventing these artifacts from propagating into GarmentSewer predictions.
  }
  \label{fig:interpolation}
\end{figure}

\paragraph{Geometry Image}
To construct the geometry image, we rasterize the 3D garment surface into the packed UV layout. For each panel, mesh vertices are first mapped to their corresponding UV coordinates, and the 3D positions are written into the geometry image at those pixel locations. Since meshes are typically far sparser than the resolution of the UV grid, this direct rasterization produces incomplete regions. A key challenge in constructing the geometry image is obtaining smooth and reliable values both inside panel and along boundaries. As illustrated in Fig.~\ref{fig:interpolation}, relying solely on barycentric interpolation produces visibly jagged boundary artifacts, since the interpolation is restricted to the discrete triangulation and does not align with the true geometric contour of the panel. These irregularities become problematic during training because GarmentSewer employs an edge-sensitive regression loss with higher weights assigned to pixels near boundaries. If the training data contain jagged or discontinuous boundary signals, the model is forced to reproduce these artifacts, which degrades both reconstruction quality and stitching consistency. To mitigate this issue, we adopt a hybrid interpolation scheme: linear interpolation is used along panel boundaries to create smooth, consistent edge values, while barycentric interpolation is applied only within triangle interiors. This produces a dense and smooth geometry field that better reflects the underlying surface and avoids introducing unwanted artifacts into the learning process, see~\cref{algo:geo-image}.

\begin{algorithm}[ht]
\caption{Geometry Image Creation from Packed Layout}
\label{algo:geo-image}
\begin{algorithmic}[1]
\Require Panels $\mathbf{P}$ with mesh vertices and faces, packed layout $L_{\text{UV}}$
\Ensure Geometry image $GGI_{\text{geo}}$
\State $GGI_{\text{geo}} \gets$ blank image with the same resolution as $L_{\text{UV}}$

\For{each panel $P_i \in \mathbf{P}$}
    \State $(V_i, F_i) \gets$ vertices and faces of $P_i$
    \State $U_i \gets$ UV coordinates of $V_i$ from $L_{\text{UV}}$

    \State
    \For{each vertex $v \in V_i$ with UV $u \in U_i$} \Comment{Vertex rasterization}
        \State $GGI_\text{geo}(u) \gets v$ 
    \EndFor

    \State
    \State $E_i \gets$ boundary edges of $P_i$
    \For{each edge endpoints $(v_a, v_b) \in E_i$} \Comment{Edge interpolation}
        \State $u_a, u_b \gets$ the corresponding UV coordinates of $v_a$ and $v_b$
        \State Sample continuous UV coordinates $\{u_k\}$ along $u_a$ and $u_b$
        \For{each sampled coordinate $u_k$}
            \State $\alpha \gets \frac{\lVert u_k - u_a \rVert}{\lVert u_b - u_a \rVert}$
            \State $GGI_{\text{geo}}(u_k) \gets (1 - \alpha)\, v_a + \alpha\, v_b$ \Comment{Linear interpolation}
        \EndFor
    \EndFor

    \State
    \For{each face $(v_a, v_b, v_c) \in F_i$} \Comment{Interior barycentric interpolation}
        \State $u_a, u_b, u_c \gets$ the corresponding UV coordinates of $v_a$, $v_b$, and $v_c$
        \State Identify all UV coordinates $\{u_k\}$ inside the triangle of $(u_a, u_b, u_c)$
        \For{each coordinate $u_k$}
            \State $GGI_{\text{geo}}(u_k) \gets \text{barycentric\_interpolation}(v_a, v_b, v_c, u_a, u_b, u_c, u_k)$
        \EndFor
    \EndFor
\EndFor

\State \Return $GGI_{\text{geo}}$
\end{algorithmic}
\end{algorithm}

\subsection{Postprocessing pipeline}

\begin{figure}[h]
    \centering
  \includegraphics[width=\textwidth]{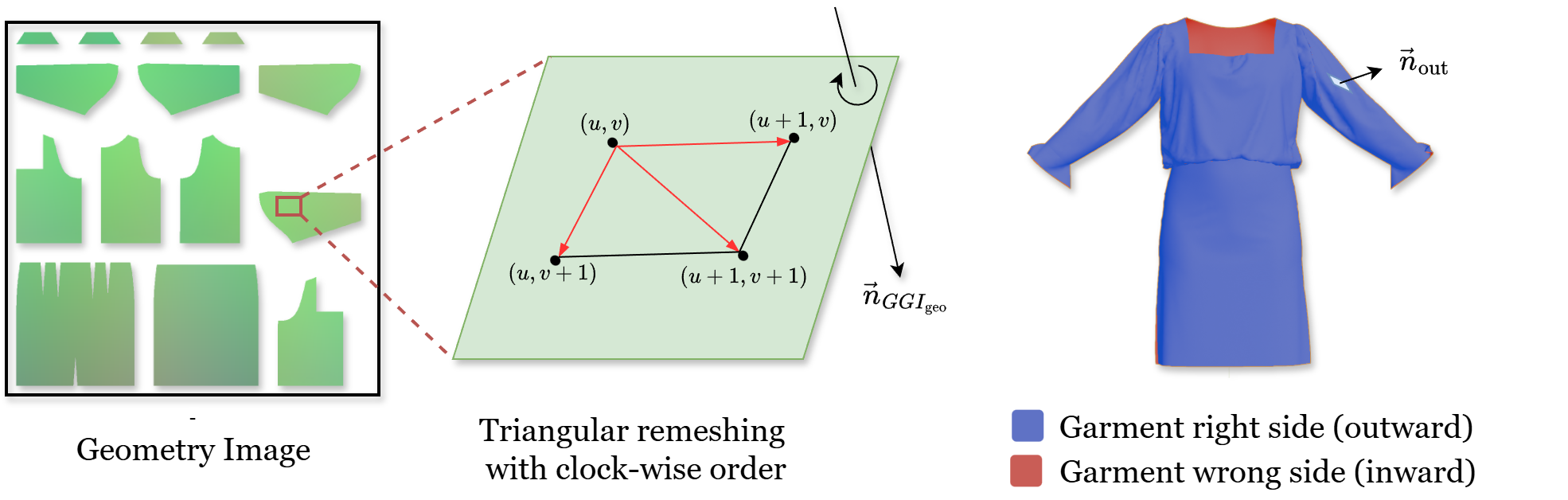}
  \caption{Remeshing from the geometry image. Starting from the UV-aligned geometry image, we perform local triangular remeshing by examining each $2 \times 2$ UV cell and generating one or two triangles depending on valid vertex occupancy. When all four vertices are present, the diagonal yielding the shorter 3D distance is selected. All faces are constructed in clockwise order to ensure consistent outward-facing normals across panels.}

  \label{fig:remeshing}
\end{figure}

\paragraph{Remeshing}
Given the predicted geometry image, the first step in the postprocessing pipeline is to recover a valid triangular mesh for each panel. As shown in~\cref{fig:remeshing}, we perform remeshing directly in UV space by scanning the geometry image in a grid-aligned manner. For every $2\times2$ UV cell, we examine the occupancy of its four pixels and generate either one or two triangles depending on how many valid vertices are present. When all four pixels contain valid geometry, we select the diagonal that yields the shorter 3D distance, ensuring a consistent and well-shaped triangulation. The pseudo-code is provided in~\cref{algo:remeshing}. All triangles are constructed with clockwise orientation so that the resulting face normals sharing the same normal with geometry image $\vec{n}_{GGI_\text{geo}}$ and follow a consistent outward normal direction (right side of the garment) $\vec{n}_\text{out}$, which is crucial for later stitching and rendering. This UV-aligned remeshing produces a dense and topologically clean mesh for each panel without needing a physics-based surface reconstruction step.

\begin{algorithm}[h]
\caption{Remeshing from Geometry Image}
\label{algo:remeshing}
\begin{algorithmic}[1]
\Require Geometry image $GGI_{\text{geo}}$
\Ensure Vertex array $V$, face array $F$, occupancy map $O$, vertex index map $I$

\For{each UV coordinate $u$}
    \State $O(u) \gets 1$ if $GGI_{\text{geo}}(u)$ contains a valid 3D vertex, else $0$
    \If{$O(u)=1$}
        \State $I(u) \gets$ assign a unique vertex index
        \State $V[I(u)] \gets GGI_{\text{geo}}(u)$
    \EndIf
\EndFor

\State $F \gets \emptyset$

\For{$x = 0$ to $H-2$}
    \For{$y = 0$ to $W-2$}
        \State $(x_0,y_0) \gets (x,y)$,\; $(x_1,y_1) \gets (x+1,y)$
        \State $(x_2,y_2) \gets (x,y+1)$,\; $(x_3,y_3) \gets (x+1,y+1)$
        \State $\mathcal{S} \gets \{(x_0,y_0),(x_1,y_1),(x_2,y_2),(x_3,y_3)\}$
        \State $\mathcal{S}_{\text{valid}} \gets \{(x',y') \in \mathcal{S} \mid O(x',y') = 1\}$

        \If{$|\mathcal{S}_{\text{valid}}| < 3$}
            \State \textbf{continue}

        \ElsIf{$|\mathcal{S}_{\text{valid}}| = 3$} \Comment{One triangle}
            \State Let $(x_a,y_a),(x_b,y_b),(x_c,y_c)$ be the three valid pixels
            \State Order $(I(x_a,y_a), I(x_b,y_b), I(x_c,y_c))$ clockwise
            \State Add the triangle to $F$

        \Else \Comment{Two triangles}
            \State $i_{00} \gets I(x_0,y_0)$,\; $i_{10} \gets I(x_1,y_1)$
            \State $i_{01} \gets I(x_2,y_2)$,\; $i_{11} \gets I(x_3,y_3)$
            \State $d_1 \gets \|V[i_{00}] - V[i_{11}]\|$
            \State $d_2 \gets \|V[i_{10}] - V[i_{01}]\|$

            \If{$d_1 \le d_2$}
                \State Add faces $(i_{00}, i_{10}, i_{11})$ and $(i_{00}, i_{11}, i_{01})$ to $F$
            \Else
                \State Add faces $(i_{00}, i_{10}, i_{01})$ and $(i_{10}, i_{11}, i_{01})$ to $F$
            \EndIf
        \EndIf
    \EndFor
\EndFor

\State \Return $(V, F, O, I)$
\end{algorithmic}
\end{algorithm}

\begin{figure}[ht]
    \centering
  \includegraphics[width=\textwidth]{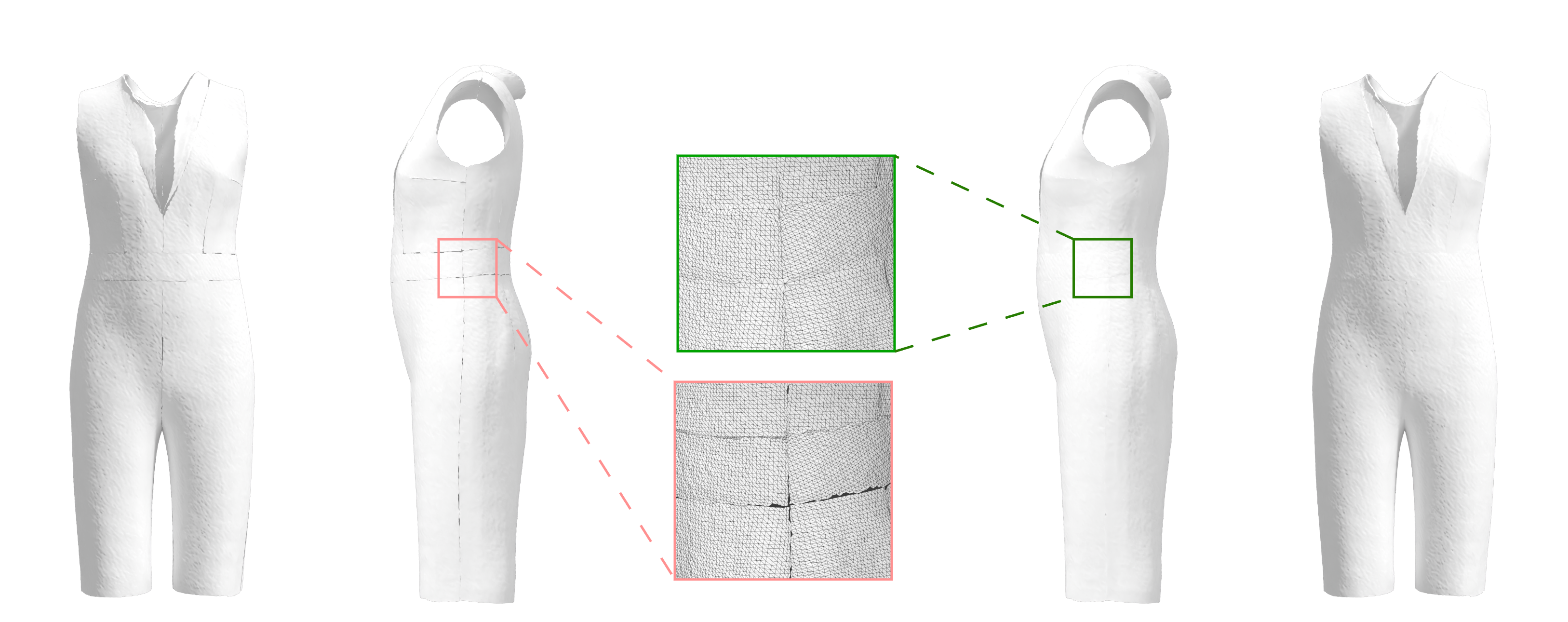}
  \caption{Stitching results before and after seam alignment. Using the stitching image, boundary edges are paired and aligned via Dynamic Time Warping, followed by vertex merging through a disjoint-set union. The zoomed-in wireframe views highlight how stitching resolves discontinuities and removes gaps between corresponding panel edges, achieving globally coherent garment mesh. The example is conducted on predicted sewing pattern from PatternMaker.}

  \label{fig:remeshing-and-stitching}
\end{figure}

\paragraph{Stitching}
After remeshing individual panels, we restore global garment connectivity using the stitching image. \cref{fig:remeshing-and-stitching} shows the resulting improvement before and after stitching, including zoomed-in wireframe views of the seam regions. Each stitched pair of edges is first extracted from the stitching image and aligned using Dynamic Time Warping to obtain a one-to-one correspondence along their UV boundary curves. The corresponding 3D vertices are then merged through a disjoint-set union, followed by averaging the vertex positions to ensure geometric consistency at the seam. Finally, degenerate faces arising from the merge are removed. This stitching step produces watertight and smoothly connected panel boundaries, removing the discontinuities that would otherwise appear in the initial, panel-wise remeshed output (see~\cref{algo:stitching}). Together with the remeshing stage, this completes the conversion from the predicted Garment Geometry Image into a coherent and simulation-ready 3D garment mesh.

\begin{algorithm}[ht]
\caption{Stitching Panels via Dynamic Time Warping and Disjoint Set Union}
\label{algo:stitching}
\begin{algorithmic}[1]
\Require Stitching image $GGI_{\text{stitch}}$, vertices $V$, faces $F$, vertex index map $I$
\Ensure Updated vertices $V$ and faces $F$

\State Extract boundary UV coordinates from $GGI_{\text{stitch}}$ and group them by stitch identifier

\State $\mathcal{E} \gets \emptyset$
\For{each stitched edge pair $(E_a, E_b)$}
    \State $\mathcal{C} \gets \text{Dynamic\_Time\_Warping}(E_a, E_b)$
    \For{each correspondence $(u_a, u_b) \in \mathcal{C}$}
        \State Add $(I(u_a), I(u_b))$ to $\mathcal{E}$
    \EndFor
\EndFor

\State Initialize DSU over all vertex indices
\For{each $(i_a, i_b) \in \mathcal{E}$}
    \State Union$(i_a, i_b)$
\EndFor

\State Merge vertices in $V$ according to DSU representatives by averaging their 3D vertex coordinates
\State Update $F$ by replacing each index with its representative and removing degenerate faces

\State \Return $V, F$
\end{algorithmic}
\end{algorithm}

\clearpage
\section{Experiments}
\label{sec:experiment}

\subsection{Experiment setup}
\paragraph{Data Setup}
We evaluate all methods on the test split of GCDMM~\cite{aipparel}, an extended version of GarmentCodeData~\cite{gcd} that includes text prompts and editing instructions. This split contains more than 5,000 samples and is used for assessing sewing pattern generation accuracy. Since computing 3D metrics and running full garment construction is significantly more expensive, we randomly extract a subset of 500 samples from test set for evaluating 3D garment generation. For SewingLDM~\cite{sewingldm} under the image-conditioned setting, we follow the authors' instructions and extract garment-only sketches from the front-view image, excluding the SMPL body from the scene.

\paragraph{Sewing Pattern Generation}
In the supplementary, we report sewing pattern generation results using image only inputs. The pattern generation metrics, including panel count accuracy, edge accuracy, and rigid transformation errors, require the model to recover precise structural details of the garment. Text inputs alone do not provide sufficiently strong geometric cues to guide any existing method toward predicting the exact panel layout or edge topology. For this reason, we exclude the text only setting from our pattern generation evaluation.

\paragraph{Garment Mesh Generation} Our goal in this stage is to measure how reliably each method can convert a predicted sewing pattern into a valid 3D garment mesh. Because predicted patterns are not always directly convertible, each pattern generator is given up to 20 attempts to produce a pattern that successfully reconstructs into a mesh. The first successful attempt is used for evaluation; if no valid reconstruction is obtained within the budget, the result is recorded as an empty point cloud. In tables of this section, we also report the average number of sampling until success.
The supplementary further presents results under single condition inputs, including text only and image only settings, to isolate the contribution of each type of conditioning information. This allows us to analyze how well each baseline and our SwiftTailor pipeline perform when restricted to a single source of information.

\subsection{Sewing Pattern Generation}

\paragraph{Image-guided Pattern Generation.} For the image-guided setting, the visual input provides strong cues about panel shapes and garment structure. As shown in~\cref{tab:pattern_prediction_image}, our method achieves the lowest geometric errors across all continuous parameters and the highest accuracy on discrete structural components. Compared to AIpparel\cite{aipparel}, ChatGarment\cite{chatgarment}, and SewingLDM\cite{sewingldm}, our model more reliably recovers the correct number of panels, edge configurations, and stitching relations from a single garment image. These improvements highlight the effectiveness of PatternMaker in leveraging image features for fine-grained structural reasoning and producing sewing patterns.

\begin{table*}[h]
\centering
\caption{Quantitative results on sewing-pattern generation with image condition only. Best results are shown in \textbf{bold}.}
\label{tab:pattern_prediction_image}
\begin{tabular}{lcccccc}
\toprule
Method & Vertex L2 ($\downarrow$) & \#Panel Acc ($\uparrow$) & \#Edge Acc ($\uparrow$) & Rot L2 ($\downarrow$) & Transl L2 ($\downarrow$) & \#Stitch Acc ($\uparrow$) \\
\midrule
AIpparel~\cite{aipparel} & 5.18 & 89.94 & 75.76 & 0.007 & 2.51 & 71.04 \\
ChatGarment~\cite{chatgarment} & 16.47 & 14.05 & 39.08 & 0.057 & 19.99 & 30.58 \\
SewingLDM~\cite{sewingldm} & 19.41 & 15.42 & 42.77 & 0.107 & 25.04 & 28.17 \\
\midrule
\textbf{Ours} & \textbf{3.70} & \textbf{91.04} & \textbf{88.96} & \textbf{0.006} & \textbf{1.91} & \textbf{83.69} \\
\bottomrule
\end{tabular}
\end{table*}


\subsection{Garment Mesh Generation}

\paragraph{Image-guided 3D Garment Generation.} Our method achieves the lowest MMD and the highest coverage under image-only conditioning in~\cref{tab:image_garment_construction}, showing that it reconstructs 3D garments that are both closer to the reference distribution and more diverse. Compared to pairing PatternMaker with GarmentCode\cite{gc}, replacing the construction stage with GarmentSewer reduces MMD (from 6.82 to 5.23) and improves COV (from 0.56 to 0.68).

\begin{table}[h]
\centering
\small
\caption{Quantitative results on mesh generation using image condition only. Best results are shown in \textbf{bold}.}
\label{tab:image_garment_construction}
\begin{tabular}{lccc}
\toprule
Method & MMD $\downarrow$ & COV $\uparrow$ & \#Sampling$\downarrow$ \\
\midrule
AIpparel\cite{aipparel} + GC\cite{gc} & 6.95 & 0.52 & 3.93 \\
ChatGarment\cite{chatgarment} + GC\cite{gc} & 11.64 & 0.22 & \textbf{1.31}\\
SewingLDM\cite{sewingldm} + GC\cite{gc} & 10.56 & 0.37 & 2.07 \\
\midrule
\textbf{PatternMaker + GC}\cite{gc} & 6.82 & 0.56 & 2.93 \\
\textbf{Ours} & \textbf{5.23} & \textbf{0.68} & 2.93\\
\bottomrule
\end{tabular}
\end{table}

\paragraph{Text-guided 3D Garment Generation.} \cref{tab:text_garment_construction} reports results for text-only conditioning with weaker geometric cues. Our approach still improves over other pipelines in MMD, while maintaining similar coverage as PatternMaker + GarmentCode. The gap between methods is smaller than in the image-only case, which aligns with the difficulty current pattern generators face when inferring precise panel geometry from textual descriptions. Nevertheless, once a plausible pattern is obtained, the proposed construction stage consistently produces reliable 3D meshes compared to physics engine.

\begin{table}[h]
\centering
\small
\caption{Quantitative results on mesh generation using text condition only. Best results are shown in \textbf{bold}.}
\label{tab:text_garment_construction}
\begin{tabular}{lccccc}
\toprule
Method & MMD $\downarrow$ & COV $\uparrow$ & \#Sampling$\downarrow$ \\
\midrule
AIpparel\cite{aipparel} + GC\cite{gc} & 8.59 & 0.39  & 3.76\\
ChatGarment\cite{chatgarment} + GC\cite{gc} & 12.89 & 0.20 & \textbf{1.15} \\
SewingLDM\cite{sewingldm} + GC\cite{gc} & 19.97 & 0.26 & 4.80 \\
\midrule
\textbf{PatternMaker + GC}\cite{gc} & 8.58 & \textbf{0.43} & 2.70\\
\textbf{Ours} & \textbf{7.80} & 0.42 & 2.70 \\
\bottomrule
\end{tabular}
\end{table}


\paragraph{Modular Exchanges.} \cref{tab:modular_exchange}  evaluates modularity by replacing GarmentCode with our GarmentSewer in the second stage. For all pattern generators except ChatGarment~\cite{chatgarment}, plugging in GarmentSewer yields a clear improvement in final mesh quality, with lower MMD and slightly higher COV, while keeping the number of sampling attempts unchanged. This demonstrates that our construction module can be seamlessly integrated into existing pipelines to enhance 3D reconstruction performance without modifying the upstream components. For ChatGarment~\cite{chatgarment}, which outputs coarse attribute-based patterns, the effect remains limited, reflecting upstream representation constraints rather than limitations of the construction stage. Pairing SewingLDM~\cite{sewingldm} with GarmentSewer also produces a notable drop in MMD compared to SewingLDM~\cite{sewingldm} + GarmentCode~\cite{gc}, and the full SwiftTailor pipeline combining PatternMaker with GarmentSewer achieves the best overall balance of MMD, COV, and sampling cost among all combinations.

\begin{table}[h]
\centering
\small
\caption{Quantitative results of all combinations between pattern generator and garment constructor on mesh generation using multi-modal inputs (image and text).}
\label{tab:modular_exchange}
\begin{tabular}{lcccc}
\toprule
Method & MMD$\downarrow$ & COV$\uparrow$ & \#Sampling$\downarrow$ \\
\midrule
AIpparel~\cite{aipparel} + GC~\cite{gc} & 6.94 & 0.52 &  4.27 \\
AIpparel~\cite{aipparel} + GarmentSewer & 6.03 & 0.63 &  4.27 \\
\midrule
ChatGarment~\cite{chatgarment} + GC~\cite{gc} & 12.27 &	0.22 &  1.20 \\
ChatGarment~\cite{chatgarment} + GarmentSewer & 12.56 & 0.23 &  1.20 \\
\midrule
SewingLDM~\cite{sewingldm} + GC~\cite{gc} & 11.33 & 0.34	&  5.87 \\
SewingLDM~\cite{sewingldm} + GarmentSewer & 10.96 & 0.41 &  5.87 \\
\midrule
\textbf{PatternMaker + GC~\cite{gc}} & 6.82 & 0.54 & 2.98 \\
\textbf{SwiftTailor (Ours)} & 5.31 & 0.68 & 2.98 \\
\bottomrule
\end{tabular}
\end{table}

\subsection{Qualitative Results}
\begin{figure}[ht]
    \centering
  \includegraphics[width=\textwidth]{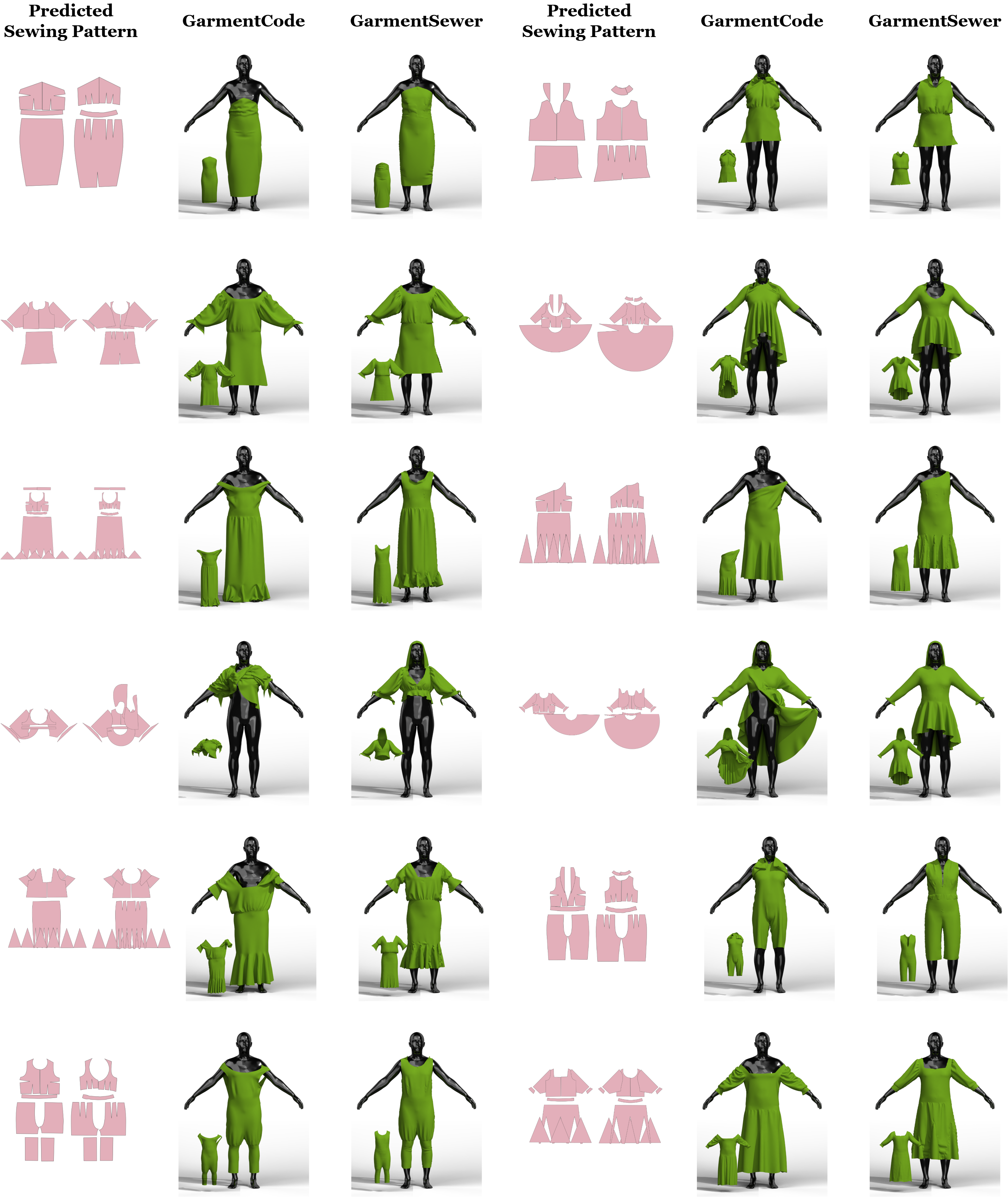}
  \caption{
    Qualitative comparison between GarmentCode~\cite{gc} and GarmentSewer given the same sewing patterns predicted by PatternMaker. GarmentSewer produces stable initializations and consistent draping, while GarmentCode\cite{gc} often fails due to rule-based panel placement.
  }
  \label{fig:gc-and-gs}
\end{figure}

\paragraph{Qualitative Comparison between GarmentSewer and GarmentCode.} \cref{fig:gc-and-gs} illustrates the differences between GarmentCode\cite{gc} and GarmentSewer when reconstructing a 3D garment from the same predicted sewing pattern produced by PatternMaker. GarmentCode\cite{gc} follows a rule-based pipeline that places 2D panels around the body and stitches them through a physics-driven sewing process. This initialization often leads to unfavorable starting states, such as panels intersecting the body or being arranged with incorrect relative orientation. As a result, the subsequent simulation struggles to recover a stable configuration, which can produce collapsed folds, tangled regions, or unrealistic draping.

In contrast, GarmentSewer directly reconstructs a geometry-image representation of the final garment silhouette, providing a stable and coherent 3D initialization before refinement. Because the mesh is already globally consistent at the start, the local relaxation during post-processing only needs to resolve minor geometric adjustments rather than repairing major structural errors. This allows GarmentSewer to preserve the intended panel relationships more faithfully and produce garments with cleaner silhouettes, smoother draping, and more reliable seam alignment. The qualitative differences across diverse garment types in~\cref{fig:gc-and-gs} highlight that GarmentSewer can avoid the failure modes seen in GarmentCode~\cite{gc}, especially in cases involving asymmetric patterns or complex multi-panel structures.

\begin{figure}[ht]
    \centering
  \includegraphics[width=0.8\textwidth]{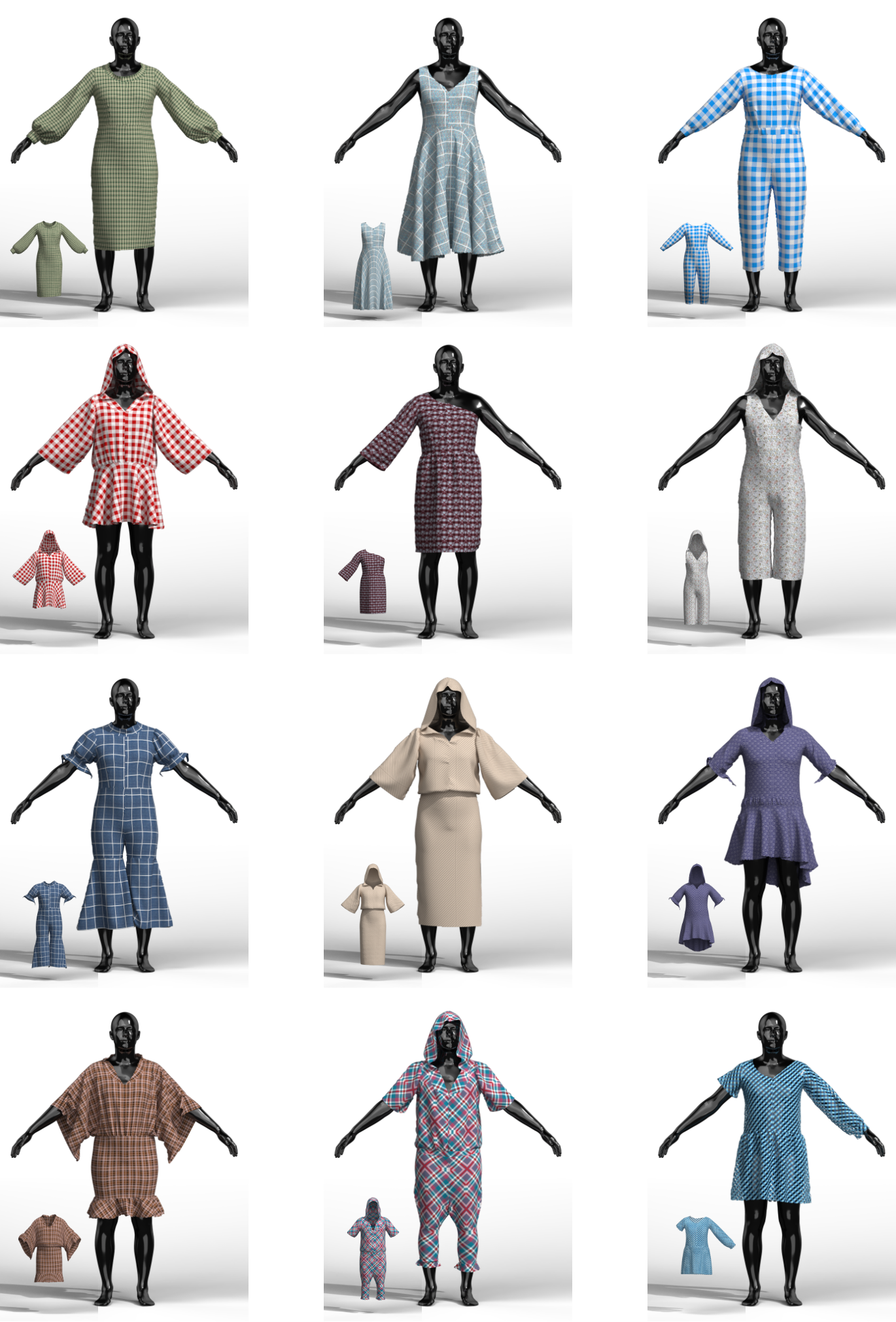}
  \caption{
    Additional qualitative results from our pipeline. Each example shows the re-draped garment on the SMPL~\cite{smpl} body together with its initial state constructed by GarmentSewer (the smaller mesh on the left). Textures are added to enhance visualization of garment geometry and structure.
  }
  \label{fig:more-qualitative}
\end{figure}

\paragraph{More qualitative results} We provide additional qualitative examples produced by our pipeline under multimodal inputs. These results are presented in~\cref{fig:more-qualitative}.

\section{Discussions}
\label{sec:discussions}
Despite the high-fidelity results and efficient inference enabled by our pipeline, several limitations remain. A key challenge is the absence of high-frequency wrinkles in the reconstructed meshes. This limitation does not arise from the geometry-image representation itself, but from the behavior of GarmentSewer during training. The model naturally learns to smooth out fine geometric variations in order to preserve global structure and ensure stable reconstruction, which results in clean and visually coherent meshes but suppresses subtle wrinkles and fold patterns. While these smooth meshes are suitable for visualization and garment showcasing, restoring realistic high-frequency wrinkles remains an open problem. One promising direction is to apply a lightweight physics-based refinement or learning-based approach to restore wrinkle on top of our stable initialization, enabling wrinkle recovery without relying on full-scale simulation.

Another limitation lies in the robustness of the pipeline under challenging, in-the-wild inputs. PatternMaker and the downstream reconstruction stages are designed around curated datasets with relatively clean observations and well-structured garments. When confronted with complex backgrounds, occlusions, unconventional silhouettes, or garments far outside the training distribution, failure cases become more frequent. Extending the system to handle broader visual variability through stronger vision encoders, data augmentation, or explicit garment parsing will improve reliability in real-world scenarios.

Finally, the modular design of our pipeline creates opportunities for future extensions. Examples include reconstructed garments with material properties or support realistic downstream simulation, or integrating user-driven editing interfaces that operate directly on predicted meshes. These directions broaden the scope from reconstruction toward interactive and controllable garment modeling.

\end{document}